\documentclass[11pt]{article}

\usepackage[final]{acl}

\usepackage{times}
\usepackage{latexsym}

\usepackage[T1]{fontenc}

\usepackage[utf8]{inputenc}

\usepackage{microtype}

\usepackage{inconsolata}

\usepackage{graphicx}
\usepackage{amsmath}
\usepackage{amssymb}
\usepackage{booktabs,multirow}
\usepackage[table]{xcolor}

\newcommand{\maincell}[1]{\cellcolor{gray!10}{#1}}
\usepackage{booktabs}
\usepackage{graphicx}
\usepackage{url}
\title{When Irrelevant Text Matters: Affine Margin Shifts in Multimodal Large Language Models}

\author{
    \textbf{Yinfeng Wang$^{1}$, Zhiyuan Yao$^{1}$, Zheren Fu$^{1}$, Lei Zhang$^{1}$, Zhendong Mao$^{1,2,}$\thanks{Corresponding author.} } \\
    $^{1}$University of Science and Technology of China \\
    $^{2}$Institute of Artificial Intelligence, Hefei Comprehensive National Science Center \\
    \texttt{\{wyf666, yaozhiyuan\}@mail.ustc.edu.cn} \\
    \texttt{\{fzr,leizh23,zdmao\}@ustc.edu.cn} \\
}

\begin{document}
\maketitle
\begin{abstract}
Multimodal large language models (MLLMs) are frequently exposed to auxiliary textual context, the impact of which on vision-language tasks remains underexplored. 
In this paper, we investigate the influence of task-irrelevant context by formulating it as a controlled intervention within a binary visual judgment framework. 
Through controlled paired comparisons with and without auxiliary textual context, we observe that irrelevant text consistently biases model predictions across diverse benchmarks. 
To move beyond performance metrics, we characterize this sensitivity through "decision margin", which is defined by the log-probability difference between candidates. 
Our analysis reveals an empirical pattern: context-conditioned margins are often well approximated by an affine transformation of their context-free counterparts.
We further interpret the fitted affine parameters as metrics for visual commitment preservation and directional answer bias. These findings provide a margin-level diagnostic view of irrelevant-context effects in MLLMs, indicating that irrelevant context does not manifest as stochastic noise, but as an estimable distortion of model preference.
We release all the code in \url{https://github.com/Wangyf1998/Irrelevant_Text_Matters}.
\end{abstract}

\begin{figure}[t]
\centering
\includegraphics[width=0.5\textwidth]{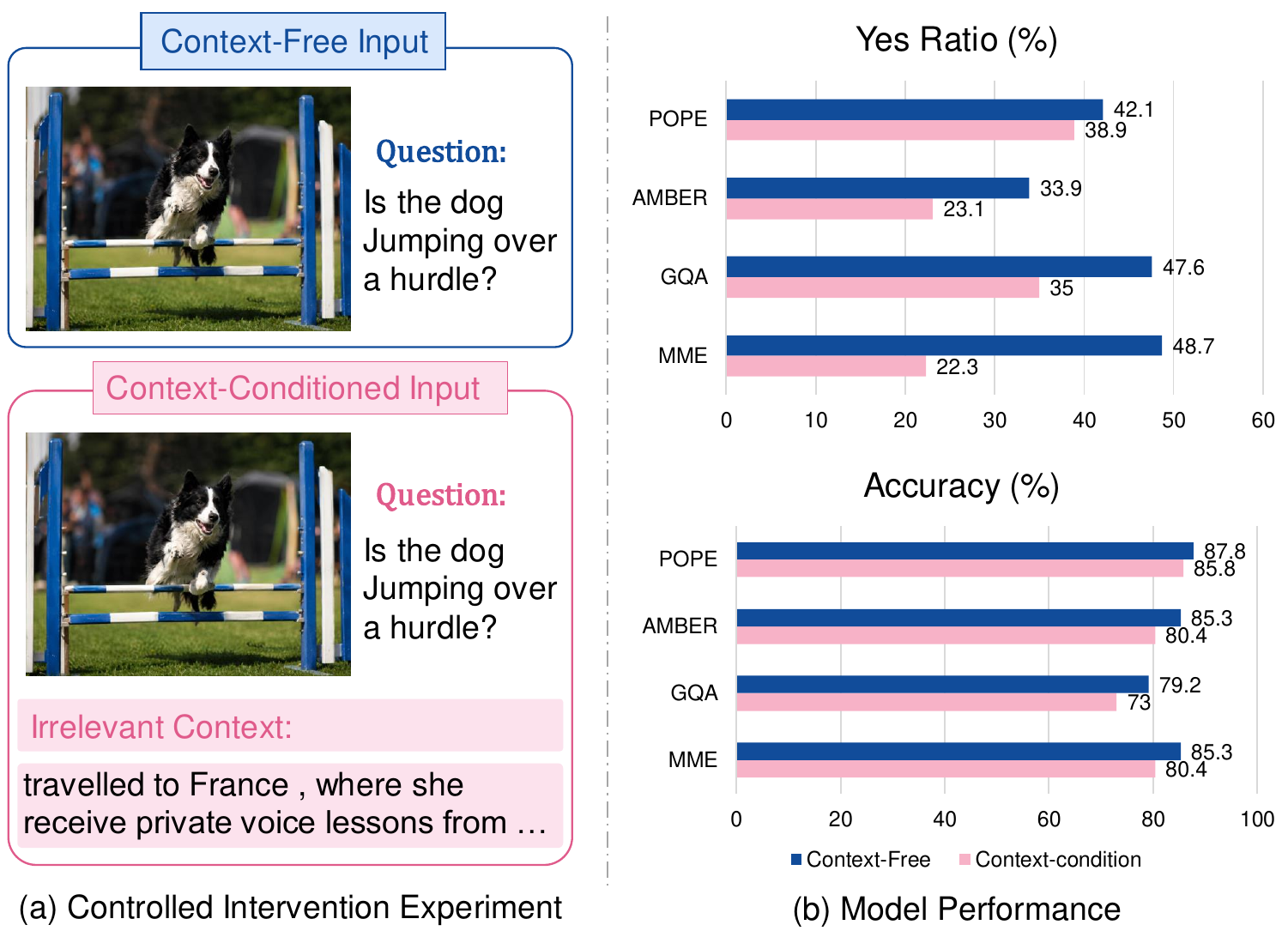}
\caption{
Overview of the controlled intervention and its behavioral effects. 
(a) For each image-question pair, we compare a context-free input with a paired context-conditioned input, where an irrelevant context is inserted. 
(b) Across multiple benchmarks, irrelevant context not only reduces predictive accuracy, but also shifts model decisions toward negative responses.
}
\label{fig:fig1}
\end{figure}

\section{Introduction}

Multimodal large language models (MLLMs) have demonstrated strong cross-modal understanding capabilities, enabling complicated vision-language tasks such as visual question answering, multi-turn reasoning, and multimodal agents. 
In these scenarios, the input image is often accompanied by auxiliary textual context, such as retrieved passages, user-provided side information, or history dialogue. 
Such context is not always aligned with the current task: retrieved passages may be off-topic~\cite{mortaheb2025reranking}, while instructions and dialogue may also contain irrelevant information~\cite{wang2024mitigating,park2024mitigating}.
This raises a question: 
\textit{Does irrelevant context affect MLLM predictions on vision-language tasks?}

Our question is motivated by a broader observation in NLP community: large language models (LLMs) can be sensitive to textual context that is not part of the task-relevant evidence. 
Prior studies have shown that irrelevant or distracting context can mislead LLMs in question answering and reasoning tasks~\cite{jia2017adversarial,shi2023large,amiraz2025distracting}. Similar issues arise in retrieval-augmented generation, where retrieved passages containing irrelevant information will affect the model's performance~\cite{yoran2024making,yan2024corrective,cuconasu2024power}.
In MLLMs, recent work shows that they may over-rely on additional textual context and resulting in a strong bias~\cite{deng2025words,chen2024mllm,zhang2024mr2ag}. 
However, despite these observations, there remains a lack of systematic investigation into how irrelevant context transforms MLLM decision-making in terms of behavioral effects and the underlying mechanisms.

\begin{figure}[t]
\centering
\includegraphics[width=0.5\textwidth]{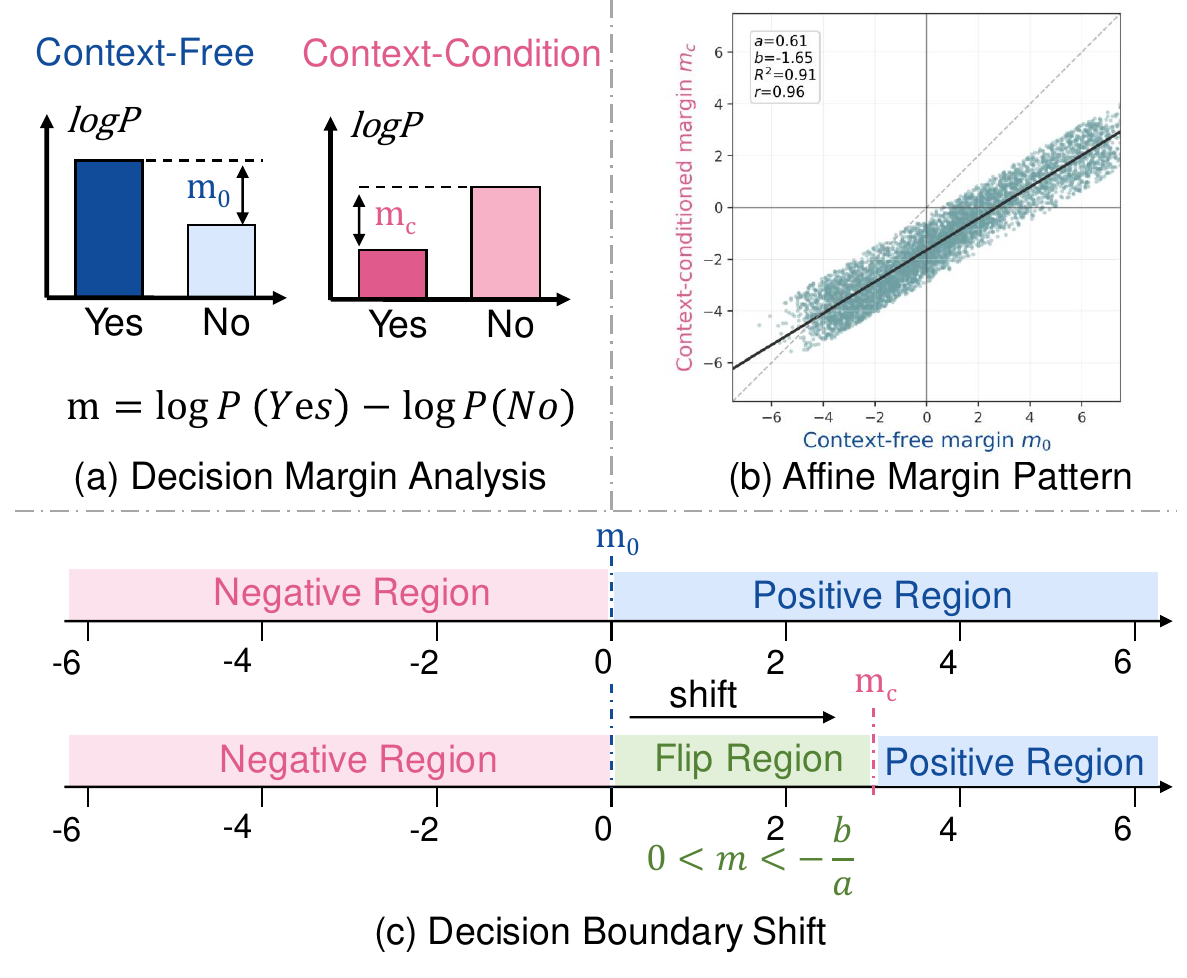}
\caption{
Illustration of the margin-level analysis.
(a) The decision margin is defined as the log-probability difference between the positive and negative answers.
(b) Adding irrelevant context transforms the context-free margin $m_0$ into the context-conditioned margin $m_c$, which follows an approximate affine relation $m_c \approx a m_0 + b$. 
(c) The affine transformation shifts the effective decision boundary from $m_0=0$ to $m_0=-b/a$. 
This shift expands the Negative Region and compresses the Positive Region. 
As a result, samples in the Flip Region, i.e., $0<m_0<-b/a$, are likely to change from positive to negative predictions, inducing a systematic shift in decision preference.
}
\label{fig:fig2}
\end{figure}

To bridge this gap, we conduct a systematic investigation into the influence of image-irrelevant text, utilizing it as a controlled intervention within binary visually grounded prediction tasks. 
Specifically, as illustrated in Figure~\ref{fig:fig1}(a), for each image-question pair, we treat the original input containing only the image and the question as the \emph{context-free} input. We then construct a paired \emph{context-condition} input by inserting an irrelevant textual passage randomly sampled from an external corpus.
As shown in Figure~\ref{fig:fig1}(b), evaluation across multiple benchmarks shows that irrelevant context not only degrades the model's overall performance, but also induces a consistent shift in decision preference toward negative responses.

To better characterize this behavior, we introduce the \emph{decision margin}, defined as the log-probability difference between the positive and negative answers, as illustrated in Figure~\ref{fig:fig2}(a). 
Surprisingly, as shown in Figure~\ref{fig:fig2}(b), the context-condition margin can be well approximated by a simple affine transformation of the context-free margin, i.e., $m_c \approx a m_0 + b$, where $m_0$ and $m_c$ denote the margins under the context-free and context-conditioned inputs, respectively.
This affine structure further explains the observed prediction flips. As illustrated in Figure~\ref{fig:fig2}(c), the effective decision boundary shifts from $m_0=0$ to $m_0=-b/a$. 
Consequently, samples with weak positive margins are likely to cross the shifted boundary and flip to negative predictions, accounting for the systematic tendency toward negative responses. 

Inspired by the implicit Bayesian interpretation of in-context learning~\cite{xie2021explanation}, we interpret the slope $a$ as the preservation of the original visual decision structure, and the intercept $b$ as the directional answer offset introduced by irrelevant text.
Based on the above observations, we further use inverse-affine calibration as a diagnostic test of whether the fitted transformation captures recoverable context-induced variation.

Our analysis extends the understanding of irrelevant-context effects in MLLMs. We show that such effects can follow an estimable affine pattern at the margin level, and provide new insight on how textual noise biases multimodal decision-making.
These findings suggest that MLLMs remain sensitive to irrelevant context, even when it provides no useful evidence. We hope our work can serve as a building block for future research on noisy-context robustness, and contribute to the development of more trustworthy and robust multimodal systems.

\section{Irrelevant Context Affects Prediction}
\label{sec:sec2}

In this section, we first introduce our experiment setup, then empirically show that irrelevant context does change MLLM predictions. 

\subsection{Problem Formulation}
\label{sec:problem_formulation}

Given an image $I$, a question $Q$, and an image-irrelevant textual context $C$, we compare model predictions under two input conditions:
\begin{equation}
x_o = (I, Q), \qquad x_c = (I, C, Q),
\end{equation}
Here, $x_o$ denotes the context-free input, and $x_c$ denotes the context-conditioned input.

We formulate each sample as a binary visual judgment. The model is asked to decide whether the visual claim implied by the question is supported by the image. We denote the binary label space as
\begin{equation}
\mathcal{Y} = \{y^+, y^-\},
\end{equation}
where $y^+$ denotes a positive judgment that the claim is supported, and $y^-$ denotes a negative judgment that the claim is not supported. 
For a model $f_\theta$, the corresponding predictions are
\begin{equation}
\hat{y}_o = f_\theta(x_o), \qquad \hat{y}_c = f_\theta(x_c),
\end{equation}
where $\hat{y}_o,\hat{y}_c \in \mathcal{Y}$.

This formulation adopts two key simplifications for controlled analysis: we cast open-ended visually grounded predictions into binary positive/negative judgments, and we treat auxiliary textual context as a controlled intervention. 
Models typically answer with "Yes" and "No" under the default instruction. 
See Appendix~\ref{app:label_verbalization} for analysis of different verbalizations, and Appendix~\ref{app:why_binary_judgment} for more discussions on problem formulation.

\begin{figure}[t]
\centering
\includegraphics[width=0.8\linewidth]{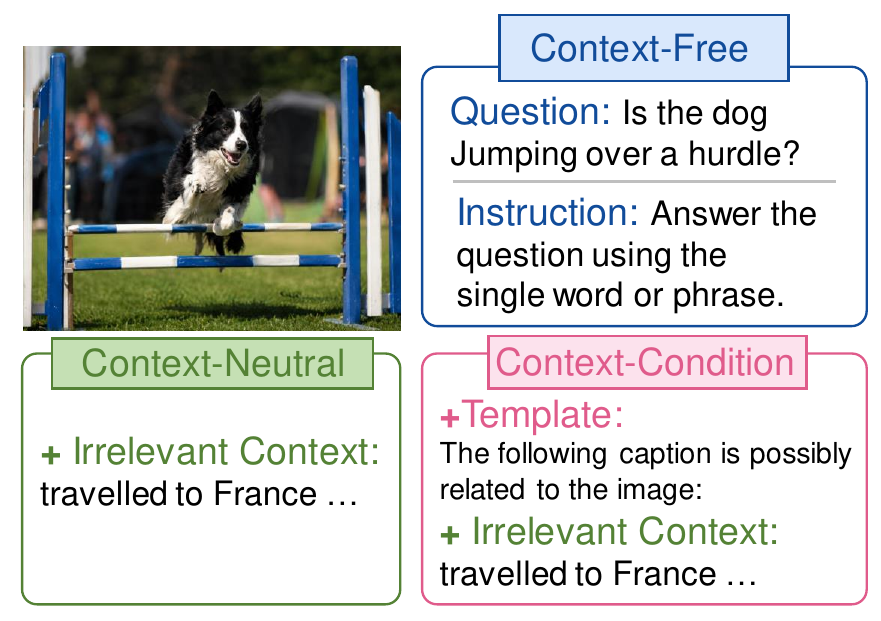}
\caption{
Visualization of three input conditions. These three conditions represent different degrees to which the irrelevant context is framed as potentially image-related.
}
\label{fig:case}
\end{figure}

\begin{table*}[t]
\centering
\small
\setlength{\tabcolsep}{2.6pt}
\caption{
Behavioral effects of image-irrelevant context across datasets and models.
Irrelevant context generally degrades performance, induces a negative-response shift, and has stronger effects when presented as potentially image-related.
\textbf{Free} indicates the context-free condition, while \textbf{$\Delta$Neu.} and \textbf{$\Delta$Con.} report changes under the context-neutral and context-condition setting relative to context-free condition.
\textbf{Yes Rate} reports the proportion of positive responses among all answers, indicating the model's tendency toward affirmative predictions.
\textbf{Flip Rate} reports the percentage of samples whose prediction changes.
All values are reported in percentage points.
}
\label{tab:table1}
\begin{tabular}{llrrrrrrrrrrr}
\toprule
\multirow{2}{*}{Dataset}
& \multirow{2}{*}{Model}
& \multicolumn{3}{c}{Accuracy}
& \multicolumn{3}{c}{Recall}
& \multicolumn{3}{c}{Yes Rate}
& \multicolumn{2}{c}{Flip Rate} \\
\cmidrule(lr){3-5}
\cmidrule(lr){6-8}
\cmidrule(lr){9-11}
\cmidrule(lr){12-13}
& & Free & $\Delta$Neu. & $\Delta$Con.
& Free & $\Delta$Neu. & $\Delta$Con.
& Free & $\Delta$Neu. & $\Delta$Con.
& Neu. & Con. \\
\midrule
\multirow{4}{*}{MME}
& LLaVA-1.5-7B
& 78.5 & $\uparrow$0.1 & \maincell{$\downarrow$0.9}
& 80.5 & $\downarrow$7.9 & \maincell{$\downarrow$8.4}
& 51.9 & $\downarrow$7.9 & \maincell{$\downarrow$7.4}
& 9.0 & \maincell{9.8} \\
& Qwen2-VL-2B
& 79.9 & $\downarrow$10.6 & \maincell{$\downarrow$20.8}
& 80.5 & $\downarrow$24.6 & \maincell{$\downarrow$40.8}
& 48.7 & $\downarrow$16.8 & \maincell{$\downarrow$26.4}
& 20.5 & \maincell{33.4} \\
& Qwen2-VL-7B
& 87.6 & $\uparrow$0.2 & \maincell{$\downarrow$27.5}
& 90.6 & $\downarrow$2.6 & \maincell{$\downarrow$69.5}
& 52.7 & $\downarrow$2.8 & \maincell{$\downarrow$41.7}
& 4.3 & \maincell{42.0} \\
& InternVL3
& 87.9 & $\downarrow$3.1 & \maincell{$\downarrow$3.7}
& 85.3 & $\downarrow$9.9 & \maincell{$\downarrow$10.5}
& 47.3 & $\downarrow$7.0 & \maincell{$\downarrow$7.0}
& 7.8 & \maincell{7.9} \\
\midrule
\multirow{4}{*}{GQA}
& LLaVA-1.5-7B
& 76.4 & $\rightarrow$0.0 & \maincell{$\downarrow$1.9}
& 76.4 & $\uparrow$2.2 & \maincell{$\uparrow$7.0}
& 50.7 & $\uparrow$2.3 & \maincell{$\uparrow$9.1}
& 7.6 & \maincell{14.3} \\
& Qwen2-VL-2B
& 79.2 & $\downarrow$6.3 & \maincell{$\downarrow$6.2}
& 76.1 & $\downarrow$20.1 & \maincell{$\downarrow$18.4}
& 47.6 & $\downarrow$14.3 & \maincell{$\downarrow$12.6}
& 16.0 & \maincell{15.6} \\
& Qwen2-VL-7B
& 82.2 & $\downarrow$1.4 & \maincell{$\downarrow$19.8}
& 78.7 & $\downarrow$2.8 & \maincell{$\downarrow$48.3}
& 47.4 & $\downarrow$1.6 & \maincell{$\downarrow$29.9}
& 3.9 & \maincell{30.0} \\
& InternVL3
& 81.9 & $\downarrow$1.5 & \maincell{$\downarrow$2.0}
& 85.1 & $\downarrow$4.3 & \maincell{$\downarrow$6.1}
& 54.2 & $\downarrow$2.9 & \maincell{$\downarrow$4.2}
& 8.0 & \maincell{9.4} \\
\midrule
\multirow{4}{*}{AMBER}
& LLaVA-1.5-7B
& 81.2 & $\downarrow$0.3 & \maincell{$\downarrow$1.3}
& 64.8 & $\downarrow$2.4 & \maincell{$\downarrow$2.3}
& 28.8 & $\downarrow$1.3 & \maincell{$\downarrow$2.9}
& 6.3 & \maincell{5.7} \\
& Qwen2-VL-2B
& 85.3 & $\downarrow$0.7 & \maincell{$\downarrow$4.9}
& 78.5 & $\downarrow$9.8 & \maincell{$\downarrow$21.6}
& 33.9 & $\downarrow$5.9 & \maincell{$\downarrow$10.8}
& 7.2 & \maincell{12.3} \\
& Qwen2-VL-7B
& 86.9 & $\uparrow$0.5 & \maincell{$\downarrow$9.4}
& 81.2 & $\downarrow$2.0 & \maincell{$\downarrow$43.1}
& 34.1 & $\downarrow$1.8 & \maincell{$\downarrow$19.7}
& 2.9 & \maincell{19.7} \\
& InternVL3
& 88.3 & $\uparrow$0.5 & \maincell{$\uparrow$0.1}
& 85.7 & $\downarrow$4.7 & \maincell{$\downarrow$6.0}
& 35.8 & $\downarrow$3.7 & \maincell{$\downarrow$4.2}
& 5.6 & \maincell{6.0} \\
\midrule
\multirow{4}{*}{POPE}
& LLaVA-1.5-7B
& 85.5 & $\downarrow$0.7 & \maincell{$\downarrow$0.3}
& 76.6 & $\downarrow$2.4 & \maincell{$\downarrow$0.9}
& 41.4 & $\downarrow$1.8 & \maincell{$\downarrow$0.6}
& 2.3 & \maincell{2.3} \\
& Qwen2-VL-2B
& 87.8 & $\downarrow$1.8 & \maincell{$\downarrow$2.0}
& 79.6 & $\downarrow$4.5 & \maincell{$\downarrow$5.1}
& 42.1 & $\downarrow$2.8 & \maincell{$\downarrow$3.2}
& 3.3 & \maincell{3.6} \\
& Qwen2-VL-7B
& 88.3 & $\downarrow$1.3 & \maincell{$\downarrow$8.4}
& 81.1 & $\downarrow$3.4 & \maincell{$\downarrow$19.8}
& 43.1 & $\downarrow$2.1 & \maincell{$\downarrow$11.6}
& 2.6 & \maincell{11.6} \\
& InternVL3
& 90.9 & $\downarrow$0.8 & \maincell{$\downarrow$0.3}
& 90.7 & $\downarrow$5.3 & \maincell{$\downarrow$3.9}
& 50.2 & $\downarrow$4.5 & \maincell{$\downarrow$3.6}
& 4.8 & \maincell{4.1} \\
\bottomrule
\end{tabular}
\end{table*}

\begin{figure*}[t]
\centering
\includegraphics[width=\textwidth]{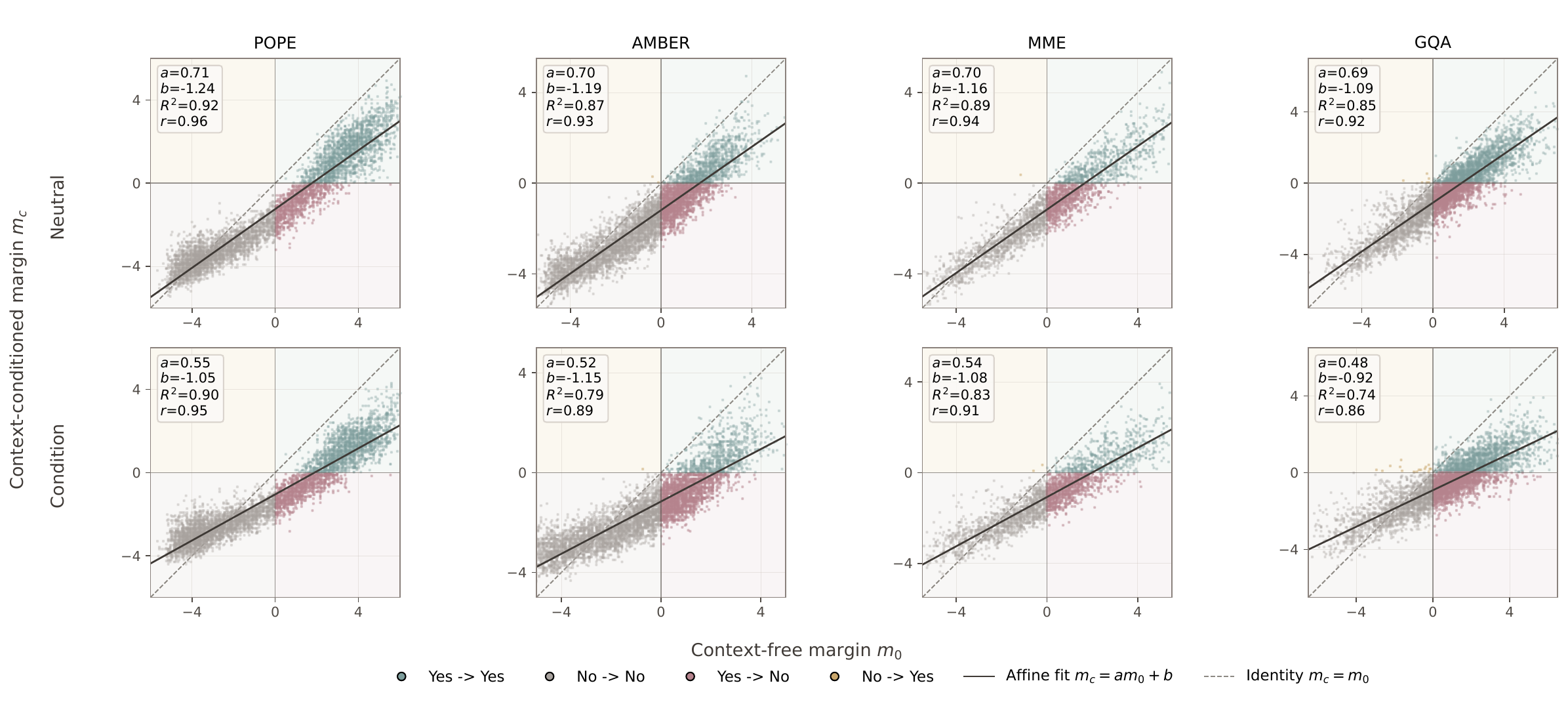}
\caption{
Affine margin patterns for LLaVA-1.5-7B across four benchmarks. 
Each panel plots the context-free margin $m_o$ against the context-conditioned margin $m_c$. 
Points are colored by prediction transitions between the two conditions, and the solid line shows the fitted affine relation $m_c = a m_o + b$; the dashed line denotes the identity relation $m_c=m_o$. 
The consistently high $R^2$ and correlation values indicate that textual context induces an approximately affine transformation of the original decision margin rather than unstructured perturbations.
Notably, changing the semantic framing from Neutral to Conditioned substantially decreases the fitted slope $a$, while leaving the negative intercept $b$ largely stable.
Results are shown for LLaVA.
}
\label{fig:affine_scatter}
\end{figure*}

\subsection{Experiment Setup}
\label{sec:experiment_setup}

\paragraph{Irrelevant Context Sampling}
Given an image-question pair $(I,Q)$, we first sample irrelevant contexts $C$ from WikiText~\cite{merity2017pointer} with a fixed random seed.
We control the length of each sampled sentence within a predefined range, and apply filtering process to ensure that the selected context is unrelated to the image. 
See Appendix~\ref{app:irrelevant_text_sampling} for implementation details.

\paragraph{Input Construction}
After irrelevant context sampling, we construct inputs under three conditions.
A case is shown in Figure~\ref{fig:case}.
For all settings, the question is followed by the same answer instruction:
\begin{quote}
\small
\ttfamily
Answer the question using a single word or phrase.
\end{quote}

The \textbf{context-free} input contains only the image and the question:
\begin{quote}
\small
\ttfamily
[Image] [Question]
\end{quote}

The \textbf{context-neutral} input appends the sampled irrelevant context:
\begin{quote}
\small
\ttfamily
[Image] [Question] [Context]
\end{quote}

The \textbf{context-conditioned} input further introduces a relevance template as:
\begin{quote}
\small
\ttfamily
The following caption is possibly related to the image:
\end{quote}
The template is placed before the context:
\begin{quote}
\small
\ttfamily
[Image] [Question] [Template] [Context]
\end{quote}

These three conditions represent different degrees to which the irrelevant text is framed as potentially image-related.
All the samples are then fed into models, and use greedy decoding to get the final answer.

\paragraph{Evaluation Benchmarks and Models}
We evaluate on four widely used vision-language benchmarks that cover diverse forms of visually grounded judgment. 
\textbf{POPE}~\cite{li2023evaluating} focuses on object-existence verification and is commonly used to evaluate object hallucination in MLLMs. 
\textbf{AMBER}~\cite{wang2023amber} extends this evaluation to multiple hallucination-related dimensions, including object existence, attributes, and relations.
\textbf{MME}~\cite{fu2023mme} provides a broader diagnostic benchmark, covering perception-oriented tasks such as color, count, OCR, and position recognition, as well as cognition-oriented tasks such as commonsense and code reasoning. 
\textbf{GQA}~\cite{hudson2019gqa} evaluates compositional visual reasoning over objects, attributes, relations, and spatial layouts. 
See Appendix~\ref{app:dataset_statistics} for data statistics and implementation detail.

For models, we report results on \textbf{LLaVA-1.5-7b}~\cite{liu2023visual}, \textbf{Qwen2-VL}~\cite{wang2024qwen2} family (2b and 7b) and \textbf{Intern-VL3-8b}~\cite{zhu2025internvl3}.

\paragraph{Discussion of Confounds}
We further examine whether the observed behavioral shift can be explained by simple experimental confounds. 
Firstly, the effect is not tied to a particular context corpus: replacing WikiText with alternative context sources, such as COCO captions~\cite{chen2015microsoft} or shuffled WikiText, preserves the main trend, although image-like captions tend to induce stronger interference. 
Secondly, it is also not explained by context length alone, since the affine distortion remains observable across different length buckets. 
Finally, prompt formatting affects the magnitude but not the existence of the phenomenon.
Detailed settings and results are provided in Appendix~\ref{app:prompt_ablation}.

\subsection{Experimental Results}
\label{sec:behavioral_results}

We report \textbf{Accuracy}, \textbf{Recall}, and \textbf{Yes rate} under the Context-free condition, together with their changes under the context-neutral and context-condition settings. We further report \textbf{Flip rate} to measure sample-level prediction changes. 

The results in Table~\ref{tab:table1} reveal three main patterns:

\paragraph{Adding irrelevant context hurts performance.}
Across models and benchmarks, adding image-irrelevant context generally reduces predictive performance. This degradation appears under both context-neutral and context-conditioned settings, indicating that auxiliary text can interfere with visually grounded prediction.

\paragraph{The behavioral shift is directional.}
The performance drop is accompanied by a systematic change in answer preference. In most settings, the Yes rate decreases after irrelevant context is added, indicating that the induced errors are not merely random.

\paragraph{Semantic framing amplifies the effect.}
The context-conditioned setting usually produces larger changes than the context-neutral setting, especially in Yes rate and flip rate. This suggests that models are more affected when the same irrelevant text is framed as potentially image-related.

These behavioral patterns suggest that irrelevant context does not act as unstructured noise; instead, it systematically changes the model's decision preferences. In the next section, we investigate this effect in detail through a margin-level analysis. See Appendix~\ref{app:absolute_behavioral_metrics} for full experiment result.

\section{Affine Margin Shift}
\label{sec:margin_analysis}

In this section, we analyze the effect of irrelevant context in logit space, and identify a simple regularity: its influence can be approximated by an affine transformation.

\subsection{Decision Margin Design}
\label{sec:decision_margin}

Given an input $x$, we define the \textit{decision margin} as the log-probability difference between the two candidate answers:
\begin{equation}
\label{equ:decision_margin}
   m(x) = \log P(y^+ \mid x) - \log P(y^- \mid x),
\end{equation}

Our design is motivated by DPO~\cite{rafailov2023direct}:
its sign indicates the model preference, while its magnitude reflects the confidence.
See Appendix~\ref{app:decision_margin_details} for implementation details.

\subsection{Affine Regularity of Decision Margins}
\label{sec:affine_relation}

Figure~\ref{fig:affine_scatter} shows the affine margin patterns for LLaVA-1.5-7B across four benchmarks. 
We identify a regularity: the impact of irrelevant text can be characterized by an affine transformation:
\begin{equation}
m_c \approx a \cdot m_0 + b,
\label{eq:affine_margin}
\end{equation}
where $a$ and $b$ represent slope and intercept estimated via least squares. To step further, we have following observations:

\begin{table}[t]
\centering
\small
\caption{
Fitted affine parameters across models and benchmarks.
The affine relationship exhibits consistently goodness of fit.
Full results of all models are shown in Appendix~\ref{app:affine_transformation_results}.
}
\label{tab:affine_parameters}
\begin{tabular}{llrrrr}
\toprule
Model & Dataset & $a$ & $b$ & $R^2$ & $r$ \\
\midrule
\multirow{4}{*}{LLaVA-1.5-7B}
& POPE       & 0.55 & -1.05 & 0.90 & 0.95 \\
& AMBER      & 0.52 & -1.15 & 0.79 & 0.89 \\
& MME        & 0.54 & -1.08 & 0.83 & 0.91 \\
& GQA        & 0.48 & -0.92 & 0.74 & 0.86 \\
\midrule
\multirow{4}{*}{Qwen2-VL-2B}
& POPE       & 0.56 & -1.58 & 0.86 & 0.93 \\
& AMBER      & 0.59 & -1.61 & 0.86 & 0.93 \\
& MME        & 0.56 & -1.13 & 0.74 & 0.86 \\
& GQA        & 0.62 & -1.55 & 0.78 & 0.88 \\
\bottomrule
\end{tabular}
\end{table}

\paragraph{The affine form is consistent across different MLLMs.}
Table~\ref{tab:affine_parameters} shows the parameters under context-neutral settings across different benchmarks and models. 
The consistently high $R^2$ values indicate that the affine relation provides a strong first-order characterization of context-induced margin shifts across the evaluated settings. Its recurrence across benchmarks suggests that this pattern is not specific to a single task distribution.

\paragraph{The affine parameters show cross-benchmark consistency.}
For a model, the fitted slope $a$ and intercept $b$ remain close across different benchmarks. Although the fitted parameters can vary with the model, task distribution and context type, it still suggests that the affine pattern reflects a model-specific response to auxiliary context.

\paragraph{Different semantic framings affect the scaling component.}
Compared with the Neutral, the Conditioned setting consistently yields a smaller slope $a$, indicating stronger margin compression. 
In contrast, the intercept term $b$ remains negative and of comparable magnitude across framings, suggesting that the answer-direction shift is relatively stable. We will provide an explanation for this phenomenon in the following section.
See Appendix~\ref{app:affine_transformation_results} and Appendix~\ref{app:robustness_analysis} for more results and robust verification.

\subsection{Boundary Shift Explains Prediction Flips}
\label{sec:boundary_shift}

Above results also explain the observed behavioral changes. 
Without context, the decision boundary is located at $m_0=0$. 
After adding context, the margin is approximately transformed as $m_c \approx a m_0 + b$, so the context-condition decision boundary satisfies $a m_0 + b = 0$.
Mapped back to the original margin axis, the effective boundary becomes
\begin{equation}
m_0 = -\frac{b}{a},
\label{eq:effective_boundary}
\end{equation}
Thus, an originally affirmative sample must have a sufficiently large positive margin to remain affirmative after irrelevant context is added;
samples with weak margins can otherwise fall below the shifted boundary and flip.

\begin{figure}[t]
\centering
\includegraphics[width=0.5\textwidth]{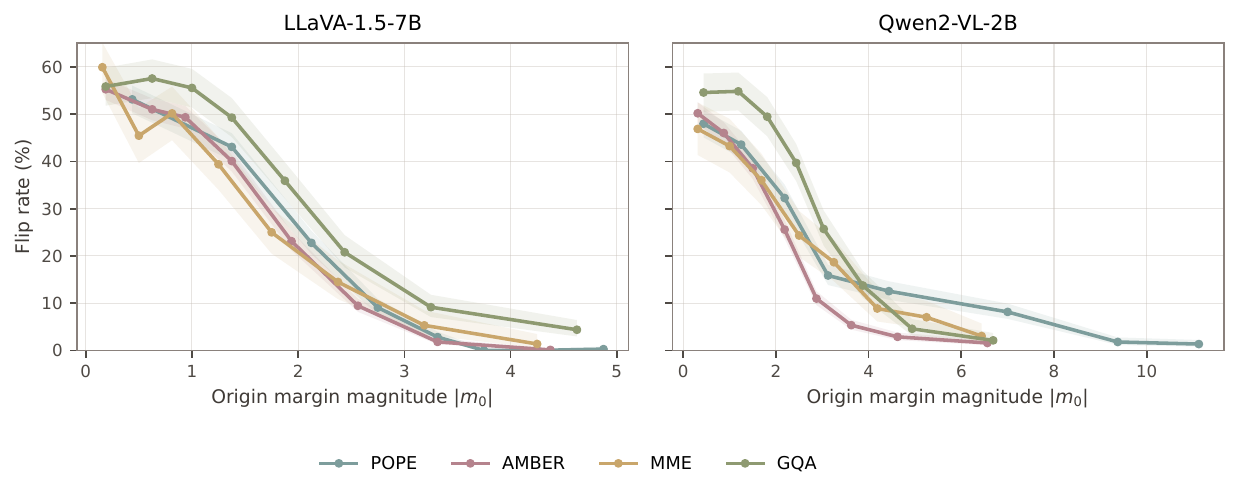}
\caption{
Flip rate across context-free margin bins.
Samples with smaller $|m_0|$ are much more likely to change their predictions after adding irrelevant context.
This indicates that context-induced flips are concentrated near the original decision boundary.
}
\label{fig:flip}
\end{figure}

We verify this explanation by grouping samples according to the magnitude of their context-free margin $|m_0|$ and computing the flip rate within each bin. 
As shown in Figure~\ref{fig:flip}, flip rates are consistently highest for low-margin samples and decrease as $|m_0|$ grows across models and benchmarks. 
This confirms that context-induced flips are not uniformly distributed over samples, but are concentrated near the original decision boundary.

\subsection{Why an Affine Approximation Can Emerge?}
The above results show that context-induced margin shifts can be well characterized by an affine relation in most settings. In this subsection, we provide a simple first-order account. 
Let $\mathcal{O}$ denote the original input, and $\mathcal{C}$ the introduced context input.
For a fixed attention query, let $s_i$ denote the pre-softmax attention score assigned to token $i$, the corresponding attention partition masses of the original input and the introduced context are defined as:

\begin{equation}
Z_{\mathcal O}=\sum_{i\in\mathcal O}\exp(s_i),\qquad
Z_{\mathcal C}=\sum_{j\in\mathcal C}\exp(s_j),
\end{equation}
After introducing the context, the attention contribution of the original tokens is subject to competition induced by normalization, with an effective scaling factor:

\begin{equation}
\lambda=\frac{Z_{\mathcal O}}{Z_{\mathcal O}+Z_{\mathcal C}},\qquad 0<\lambda<1,
\end{equation}

Let $o_0$ and $o_{\mathcal C}$ denote the normalized attention outputs aggregated from the original and context tokens, respectively, and let
$o_c$ denote the attention output after introducing the context. Assuming that the attention scores of the original tokens remain
approximately stable, softmax normalization gives:

\begin{equation}
o_c\approx\lambda o_0+(1-\lambda)o_{\mathcal C},
\end{equation}



If the downstream mapping from the resulting representation to the decision margin can be locally approximated to first order, we obtain:

\begin{equation}
m_c\approx\lambda m_0+(1-\lambda)m_{\mathcal C}=am_0+b,
\end{equation}

where $a=\lambda$ and $b=(1-\lambda)m_{\mathcal C}$. Thus, attention normalization naturally induces a multiplicative attenuation of the original margin together with an additive contextual contribution, yielding a first-order affine structure.


\subsection{Beyond binary judgments}
To examine whether the observed affine tendency is specific to controlled binary decisions, we further evaluate Multi Choice Question Answering (MCQA) and more realistic tasks, including multimodal QA, RAG, and multi-turn dialogue. The affine tendency remains observable across these settings: representative $R^2$ values range from $0.73$--$0.94$ on MCQA, $0.58$--$0.92$ for irrelevant context in realistic QA, $0.86$--$0.97$ in RAG, and $0.75$--$0.85$ with mismatched dialogue history. These results suggest that the phenomenon extends beyond binary judgments, while its fit quality varies across models, tasks, and context types. Full results and experiment settings are provided in Appendix~\ref{app:mcqa}.



\section{Interpreting the Affine Parameters}
\label{sec:boundary_crossing_analysis}

In this section, we use a lightweight Bayesian view to interpret the fitted parameters. 

\subsection{Preliminary}
\label{sec:bayesian_preliminary}

We first introduce a lightweight Bayesian-inspired view to describe how irrelevant context may enter the model's prediction process. 

Let $\mathcal{X}$ denote the original input without context. 
The model defines a context-free predictive distribution $p(Y\mid \mathcal{X})$. 
When an additional context $C$ is provided, the model may use it as an additional conditioning signal when interpreting the input. 
Following the prior work~\cite{xie2021explanation}, we introduce a latent interpretation variable $Z\in\mathcal{Z}$, which represents a task-specific “concept”. 
The context-conditioned prediction can then be written as posterior-weighted marginalization:
\begin{equation}
p(Y\mid \mathcal{X},C)
=
\int_{\mathcal{Z}}
p(Y\mid \mathcal{X},Z)\,
p(Z\mid \mathcal{X},C)\,dZ,
\label{eq:posterior_predictive_context}
\end{equation}

Under this view, adding $C$ updates the model's belief over latent interpretations:
\begin{equation}
p(Z\mid \mathcal{X},C)
\propto
p(C\mid Z,\mathcal{X})p(Z\mid \mathcal{X}),
\label{eq:context_posterior_shift}
\end{equation}
In our experiments, $C$ provides no useful evidence, i.e., $Y \perp C \mid \mathcal{X}$. 
However, image-text alignment priors may lead the model to treat $C$ as useful evidence rather than irrelevant noise. 
Consequently, $p(C\mid Z,\mathcal{X})$ may be non-uniform across latent interpretations, shifting $p(Z\mid \mathcal{X},C)$ away from the original belief $p(Z\mid \mathcal{X})$.

\subsection{Explanation of the Affine Parameters}
\label{sec:interpreting_affine_parameters}

The Bayesian formulation provides a plausible lens. 
Under this interpretation, the fitted affine map summarizes two observable effects: how much of the original visual decision structure is preserved, and whether the added context introduces a directional answer offset.

\paragraph{Slope as commitment preservation.}
We interpret the slope $a$ as the degree to which the original visual decision structure is preserved after context-conditioned inference. 
A larger $a$ indicates that the original margin geometry is better preserved, while a smaller $a$ indicates stronger attenuation of the original visual commitment.

\paragraph{Intercept as answer-direction offset.}
The intercept $b$ captures the directional offset introduced by the added context. 
Since the margin is defined such that positive values favor \texttt{Yes}, a negative $b$ corresponds to a shift toward \texttt{No}. 
For image-irrelevant context, this term is especially informative because the added text does not provide label-conditioned visual evidence and therefore tends to appear as a global directional bias in the answer space.

This interpretation explains the effect of semantic framing observed in Sec.~\ref{sec:affine_relation}.
The Neutral and Condition settings contain the same type of image-irrelevant text, but the Condition template frames the text as potentially related to the image:
this framing increases the likelihood that the model treats the irrelevant context as useful contextual evidence, thereby reweighting the latent interpretation more strongly and weakening the preservation of the original visual margin. 
At the same time, because the context is unchanged, the induced answer-direction prior is not fundamentally changed. 
However, these quantities should be interpreted as setting-dependent descriptors rather than invariant model properties.

\begin{figure}[t]
\centering
\includegraphics[width=\linewidth]{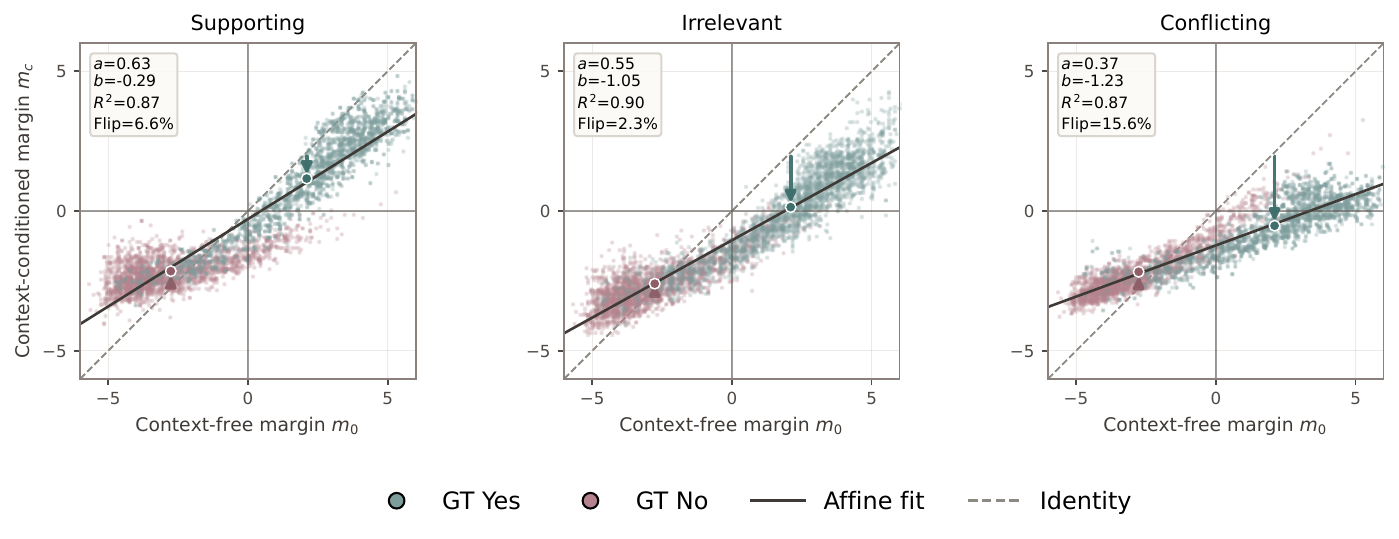}
\caption{
Experiment results for Validating the Parameter Interpretation. Points are colored by ground-truth label.
Arrows show the mean label-conditioned margin shift from the identity line to the observed context-conditioned mean.
The monotonic trend $a_{\mathrm{sup}} > a_{\mathrm{irr}} > a_{\mathrm{conf}}$ is consistent with the interpretation of $a$ as commitment preservation.
}
\label{fig:context_role_intervention}
\end{figure}

\subsection{Validating the Parameter Interpretation}
\label{sec:context_type_intervention}

To test the above implications, we construct a controlled context-type intervention. 

\paragraph{Intervention setup}

Our key idea is that, if $a$ and $b$ respectively capture visual-decision preservation and directional answer offset, then varying the evidential role of the context should change the affine parameters. Therefore, we construct three context types:
\textbf{Supporting context} is consistent with the ground-truth answer, stating that the queried object is present for positive samples and absent for negative samples.
\textbf{Irrelevant context} is the same as in the context-conditioned setting. 
\textbf{Conflicting context} states absence for positive samples and presence for negative samples. 
All context types are inserted using the same template and position in the input. 

\paragraph{Results}
Figure~\ref{fig:context_role_intervention} reports the fitted affine parameters under the three context types. 
The slope changes monotonically with the evidential role of the context:
\begin{equation}
    a_{\mathrm{sup}} > a_{\mathrm{irr}} > a_{\mathrm{conf}},
\end{equation}
This ordering is consistent with the interpretation of $a$ as commitment preservation. 
Supporting context preserves the original margin structure the most, irrelevant context introduces stronger attenuation, and conflicting context weakens the original structure most substantially. 

However, although supporting context is related, its slope still remains below one. 
We interpret this behavior as evidence that MLLMs integrate supporting context as imperfect evidence. 
See Appendix~\ref{app:supporting_slope_explanation} for more discussion.

\section{Post-Hoc Calibration Test}
\label{sec:posthoc_calibration}

In this section, we use post-hoc calibration as a diagnostic test.

\subsection{Calibration Protocol}
\label{sec:calibration_method}

Given a held-out calibration split $\mathcal{D}_{\mathrm{cal}}$, we compute the context-free margin $m_o$ and the context-conditioned margin $m_c$ for each calibration sample, and fit the affine relation.
On the test split, we apply the inverse transformation:
\begin{equation}
\label{equ:affine_calib}
    m_{\mathrm{cal}}
    =
    \frac{m_c - \hat{b}}{\hat{a}}.
\end{equation}

\subsection{Experiment Setup and results}

\paragraph{Calibration settings}
We consider two calibration settings. In the in-domain setting, affine parameters are fitted on a calibration split and applied to the held-out split of the same benchmark.

In the cross-dataset transfer setting, affine parameters are fitted on one source dataset and directly applied to other target datasets.
This setting uses no target-dataset calibration samples and tests whether the affine parameters capture a transferable model-level response to irrelevant context. 

\paragraph{Experiment Results}

Table~\ref{tab:calibration_in_domain_main} reports results of in-domain calibration, and Table~\ref{tab:calibration_transfer_main} reports the cross-dataset transfer result. Both experiments are implemented on LLaVA.
These results show that the context-induced margin distortion is structured and partially transferable: inverse-affine correction recovers context-free decisions on held-out data and under cross-dataset transfer. See Appendix~\ref{app:calibration} for more experimental details and results.

\begin{table}[t]
\centering
\small
\caption{
In-domain post-hoc affine calibration under WikiText-after-possibly context.
Affine parameters are fitted on calibration splits and evaluated on held-out
test splits. We report the Accuracy and Flip rate with percentage.
}
\label{tab:calibration_in_domain_main}
\begin{tabular}{lrrrrr}
\toprule
Dataset & Origin & Con. & Cal. & Con. Flip & Cal. Flip \\
\midrule
POPE  & 86.1 & 78.0 & 84.6 & 16.6 &  6.1 \\
AMBER & 82.7 & 74.9 & 79.7 & 29.2 & 13.6 \\
MME   & 78.7 & 68.5 & 76.5 & 30.2 & 12.6 \\
GQA   & 73.2 & 69.2 & 72.3 & 36.3 & 14.8 \\
\bottomrule
\end{tabular}
\end{table}

\begin{table}[t]
\centering
\small
\caption{
Cross-dataset affine calibration transfer result. Affine parameters are fitted on POPE and
directly applied to other benchmarks without target calibration samples.
We report the Accuracy and Flip rate with percentage.
}
\label{tab:calibration_transfer_main}
\begin{tabular}{lrrrrr}
\toprule
Dataset & Origin & Con. & Cal. & Con. Flip & Cal. Flip \\
\midrule
AMBER & 82.7 & 75.0 & 80.1 & 29.1 & 14.7 \\
MME   & 78.7 & 68.5 & 76.6 & 30.4 & 12.8 \\
GQA   & 73.2 & 69.3 & 71.7 & 36.2 & 13.1 \\
\bottomrule
\end{tabular}
\end{table}

\section{Related Work}
\label{sec:related_work}

\subsection{Context Effects in LLMs}

LLMs are known to be sensitive to contextual information that is not part of the task-relevant evidence. 
Early work on adversarial reading comprehension shows that irrelevant distractor sentences can mislead question-answering models~\cite{jia2017adversarial}, and recent studies further demonstrate that large language models can be distracted by irrelevant context in reasoning and question-answering tasks~\cite{shi2023large,amiraz2025distracting}. 
Related observations also appear in retrieval-augmented generation, where retrieved passages may contain noisy, weakly related, or irrelevant information that affects the final answer~\cite{yoran2024making,yan2024corrective,cuconasu2024power}. 

\subsection{Context Effects in MLLMs}

Textual bias has long been observed in vision-language tasks. 
Early VQA studies show that models can exploit language priors and insufficient visual grounding~\cite{goyal2017making,agrawal2018dont}. 
Subsequent work further studies how to reduce such unimodal biases and force models to rely more on visual evidence~\cite{cadene2019rubi,kv2020reducing,deng2025words}. 
In multimodal RAG, prior work shows that irrelevant retrieved passages or conflicting knowledge may reduce overall performance~\cite{chen2022rich,kortukov2024studying,caffagni2024wiki}. 

\subsection{In-Context Learning and Bias Calibration}

In-context learning (ICL) shows that language models can adapt their predictions from contextual examples without parameter updates~\cite{brown2020language,liu2022makes}. 
A theoretical explanation further views ICL as implicit Bayesian inference, where the model predicts by inferring a latent concept from the observed context~\cite{xie2021explanation}.
Despite its ability, prior work indicates that prompts in ICL may introduce systematic biases, which can be mitigated by estimating and calibrating such biases at inference time~\cite{zhao2021calibrate,han2022prototypical,nie2022improving,zhou2024batch}. 
More broadly, this connects to post-hoc calibration methods such as Platt scaling and temperature scaling, which adjust model scores without updating model parameters~\cite{platt1999probabilistic,guo2017calibration}. 

\section{Conclusion}

In this paper, we study how image-irrelevant textual context affects visually grounded decisions in MLLMs. 
By treating irrelevant context as a controlled intervention, we show that it can degrade predictive performance, shift answer preferences, and induce prediction flips across benchmarks and models. 
To characterize this effect beyond aggregate metrics, we introduce a margin-level analysis based on the log-probability difference between positive and negative answers. 
Our results show that context-conditioned margins can be well approximated by an affine transformation of context-free margins, indicating that irrelevant context induces a structured distortion of model preference rather than unstructured noise. 
We further analyze the affine parameters as compact descriptors of margin preservation and directional answer shift, and develop a post-hoc calibration method as a diagnostic test, showing that the fitted structure captures recoverable context-induced variation.
Our findings provide a new perspective on irrelevant-context effects in MLLMs and may serve as a useful basis for future work on trustworthy multimodal systems.

\section{Limitations}
\label{sec:limitations}

\paragraph{Closed-form Experiment Setting.}
Our analysis is mainly conducted in visual-question settings, and therefore does not fully capture the complexity of open-ended multimodal generation. 
We provide a preliminary study in Appendix~\ref{app:open-ended}, where we observe similar context-induced effects in open-ended generation. 
When irrelevant context is introduced, we observe that the model tends to generate shorter responses and focus more on salient objects in the image. One plausible explanation is that the context-induced affine transformation suppresses marginal visual claims with weaker confidence. In this view, objects or attributes that originally lie near the decision boundary are more likely to be filtered out after the margin shift, while highly salient objects with stronger margins remain in the output. This may explain why irrelevant context reduces response coverage and makes the generation more conservative.
However, systematically characterizing whether the same affine margin pattern extends to open-ended outputs remains an important direction for future work.

\paragraph{Generalization to Realistic Settings.}
Our main analyses are conducted under controlled settings designed to isolate the effect of auxiliary textual context. Although the affine tendency remains observable in more realistic scenarios, the fitted parameters and the context-role ordering identified under controlled interventions are not fully preserved. This suggests that these quantitative regularities are sensitive to additional factors such as context relevance, linguistic form, and cross-modal interactions, and should therefore be interpreted as controlled empirical characterizations rather than invariant properties across tasks and contexts.

\paragraph{Limited theoretical explanation.}
The first-order account in Sec. 3.4 provides a structural motivation for why an affine approximation can emerge, but it does not establish a complete causal explanation of the phenomenon.
Developing a deeper theoretical and causal account of this phenomenon remains an important direction for future work.

\section{Acknowledgments}
This research is supported by the Artificial Intelligence-National Science and Technology Major Project under Grant 2023ZD0121200, 
the Postdoctoral Fellowship Program of CPSF under Grant Number GZC20260860, 
the National Natural Science Foundation of China under Grant 62336001,
and the Fundamental and Interdisciplinary Disciplines Breakthrough Plan
of the Ministry of Education of China under No.JYB2025XDXM103.


\bibliography{custom}

@inproceedings{li2023evaluating,
  title = {Evaluating Object Hallucination in Large Vision-Language Models},
  author = {Li, Yifan and Du, Yifan and Zhou, Kun and Wang, Jinpeng and Zhao, Wayne Xin and Wen, Ji-Rong},
  booktitle = {Proceedings of the 2023 Conference on Empirical Methods in Natural Language Processing},
  year = {2023}
}

@inproceedings{liu2023visual,
  title     = {Visual Instruction Tuning},
  author    = {Liu, Haotian and Li, Chunyuan and Wu, Qingyang and Lee, Yong Jae},
  booktitle = {Advances in Neural Information Processing Systems},
  year      = {2023}
}

@inproceedings{amiraz2025distracting,
  title={The distracting effect: Understanding irrelevant passages in rag},
  author={Amiraz, Chen and Cuconasu, Florin and Filice, Simone and Karnin, Zohar},
  booktitle={Proceedings of the 63rd Annual Meeting of the Association for Computational Linguistics (Volume 1: Long Papers)},
  pages={18228--18258},
  year={2025}
}

@inproceedings{cuconasu2024power,
  title={The power of noise: Redefining retrieval for rag systems},
  author={Cuconasu, Florin and Trappolini, Giovanni and Siciliano, Federico and Filice, Simone and Campagnano, Cesare and Maarek, Yoelle and Tonellotto, Nicola and Silvestri, Fabrizio},
  booktitle={Proceedings of the 47th International ACM SIGIR Conference on Research and Development in Information Retrieval},
  pages={719--729},
  year={2024}
}

@inproceedings{jia2017adversarial,
  title     = {Adversarial Examples for Evaluating Reading Comprehension Systems},
  author    = {Jia, Robin and Liang, Percy},
  booktitle = {Proceedings of the 2017 Conference on Empirical Methods in Natural Language Processing},
  pages     = {2021--2031},
  address   = {Copenhagen, Denmark},
  publisher = {Association for Computational Linguistics},
  year      = {2017},
  doi       = {10.18653/v1/D17-1215},
  url       = {https://aclanthology.org/D17-1215/}
}

@inproceedings{shi2023large,
  title     = {Large Language Models Can Be Easily Distracted by Irrelevant Context},
  author    = {Shi, Freda and Chen, Xinyun and Misra, Kanishka and Scales, Nathan and Dohan, David and Chi, Ed H. and Sch{\"a}rli, Nathanael and Zhou, Denny},
  booktitle = {Proceedings of the 40th International Conference on Machine Learning},
  pages     = {31210--31227},
  year      = {2023},
  editor    = {Krause, Andreas and Brunskill, Emma and Cho, Kyunghyun and Engelhardt, Barbara and Sabato, Sivan and Scarlett, Jonathan},
  volume    = {202},
  series    = {Proceedings of Machine Learning Research},
  publisher = {PMLR},
  url       = {https://proceedings.mlr.press/v202/shi23a.html}
}

@inproceedings{yoran2024making,
  title     = {Making Retrieval-Augmented Language Models Robust to Irrelevant Context},
  author    = {Yoran, Ori and Wolfson, Tomer and Ram, Ori and Berant, Jonathan},
  booktitle = {The Twelfth International Conference on Learning Representations},
  year      = {2024},
  url       = {https://openreview.net/forum?id=ZS4m74kZpH}
}

@article{yan2024corrective,
  title   = {Corrective Retrieval Augmented Generation},
  author  = {Yan, Shi-Qi and Gu, Jia-Chen and Zhu, Yun and Ling, Zhen-Hua},
  journal = {arXiv preprint arXiv:2401.15884},
  year    = {2024},
  url     = {https://arxiv.org/abs/2401.15884}
}

@article{kalai2025language,
  title={Why language models hallucinate},
  author={Kalai, Adam Tauman and Nachum, Ofir and Vempala, Santosh S and Zhang, Edwin},
  journal={arXiv preprint arXiv:2509.04664},
  year={2025}
}

@article{wang2023amber,
  title   = {{AMBER}: An {LLM}-free Multi-dimensional Benchmark for {MLLMs} Hallucination Evaluation},
  author  = {Wang, Junyang and Wang, Yuhang and Xu, Guohai and Zhang, Jing and Gu, Yukai and Jia, Haitao and Wang, Jiaqi and Xu, Haiyang and Yan, Ming and Zhang, Ji and Sang, Jitao},
  journal = {arXiv preprint arXiv:2311.07397},
  year    = {2023},
  url     = {https://arxiv.org/abs/2311.07397}
}

@article{fu2023mme,
  title   = {{MME}: A Comprehensive Evaluation Benchmark for Multimodal Large Language Models},
  author  = {Fu, Chaoyou and Chen, Peixian and Shen, Yunhang and Qin, Yulei and Zhang, Mengdan and Lin, Xu and Yang, Jinrui and Zheng, Xiawu and Li, Ke and Sun, Xing and Wu, Yunsheng and Ji, Rongrong},
  journal = {arXiv preprint arXiv:2306.13394},
  year    = {2023},
  url     = {https://arxiv.org/abs/2306.13394}
}

@inproceedings{hudson2019gqa,
  title     = {{GQA}: A New Dataset for Real-World Visual Reasoning and Compositional Question Answering},
  author    = {Hudson, Drew A. and Manning, Christopher D.},
  booktitle = {Proceedings of the IEEE/CVF Conference on Computer Vision and Pattern Recognition},
  pages     = {6700--6709},
  year      = {2019},
  url       = {https://openaccess.thecvf.com/content_CVPR_2019/html/Hudson_GQA_A_New_Dataset_for_Real-World_Visual_Reasoning_and_Compositional_CVPR_2019_paper.html}
}

@inproceedings{goyal2017making,
  title     = {Making the {V} in {VQA} Matter: Elevating the Role of Image Understanding in Visual Question Answering},
  author    = {Goyal, Yash and Khot, Tejas and Summers-Stay, Douglas and Batra, Dhruv and Parikh, Devi},
  booktitle = {Proceedings of the IEEE Conference on Computer Vision and Pattern Recognition},
  pages     = {6904--6913},
  year      = {2017},
  url       = {https://openaccess.thecvf.com/content_cvpr_2017/html/Goyal_Making_the_V_CVPR_2017_paper.html}
}

@inproceedings{merity2017pointer,
  title     = {Pointer Sentinel Mixture Models},
  author    = {Merity, Stephen and Xiong, Caiming and Bradbury, James and Socher, Richard},
  booktitle = {International Conference on Learning Representations},
  year      = {2017},
  url       = {https://arxiv.org/abs/1609.07843}
}

@article{chen2015microsoft,
  title   = {Microsoft {COCO} Captions: Data Collection and Evaluation Server},
  author  = {Chen, Xinlei and Fang, Hao and Lin, Tsung-Yi and Vedantam, Ramakrishna and Gupta, Saurabh and Doll{\'a}r, Piotr and Zitnick, C. Lawrence},
  journal = {arXiv preprint arXiv:1504.00325},
  year    = {2015},
  url     = {https://arxiv.org/abs/1504.00325}
}

@article{xie2021explanation,
  title={An explanation of in-context learning as implicit bayesian inference},
  author={Xie, Sang Michael and Raghunathan, Aditi and Liang, Percy and Ma, Tengyu},
  journal={arXiv preprint arXiv:2111.02080},
  year={2021}
}

@article{chen2024mllm,
  title        = {MLLM Is a Strong Reranker: Advancing Multimodal Retrieval-Augmented Generation via Knowledge-Enhanced Reranking and Noise-Injected Training},
  author       = {Chen, Zhanpeng and Xu, Chengjin and Qi, Yiyan and Guo, Jian},
  journal      = {arXiv preprint arXiv:2407.21439},
  year         = {2024}
}

@article{zhang2024mr2ag,
  title        = {mR$^2$AG: Multimodal Retrieval-Reflection-Augmented Generation for Knowledge-Based VQA},
  author       = {Zhang, Tao and Zhang, Ziqi and Ma, Zongyang and Chen, Yuxin and Qi, Zhongang and Yuan, Chunfeng and Li, Bing and Pu, Junfu and Zhao, Yuxuan and Xie, Zehua and Ma, Jin and Shan, Ying and Hu, Weiming},
  journal      = {arXiv preprint arXiv:2411.15041},
  year         = {2024}
}

@article{mortaheb2025reranking,
  title        = {Re-ranking the Context for Multimodal Retrieval Augmented Generation},
  author       = {Mortaheb, Matin and Khojastepour, Mohammad A. Amir and Chakradhar, Srimat T. and Ulukus, Sennur},
  journal      = {arXiv preprint arXiv:2501.04695},
  year         = {2025}
}

@inproceedings{caffagni2024wiki,
  title={Wiki-llava: Hierarchical retrieval-augmented generation for multimodal llms},
  author={Caffagni, Davide and Cocchi, Federico and Moratelli, Nicholas and Sarto, Sara and Cornia, Marcella and Baraldi, Lorenzo and Cucchiara, Rita},
  booktitle={Proceedings of the IEEE/CVF Conference on Computer Vision and Pattern Recognition},
  pages={1818--1826},
  year={2024}
}

@article{wang2024qwen2,
  title={Qwen2-vl: Enhancing vision-language model's perception of the world at any resolution},
  author={Wang, Peng and Bai, Shuai and Tan, Sinan and Wang, Shijie and Fan, Zhihao and Bai, Jinze and Chen, Keqin and Liu, Xuejing and Wang, Jialin and Ge, Wenbin and others},
  journal={arXiv preprint arXiv:2409.12191},
  year={2024}
}

@article{zhu2025internvl3,
  title={Internvl3: Exploring advanced training and test-time recipes for open-source multimodal models},
  author={Zhu, Jinguo and Wang, Weiyun and Chen, Zhe and Liu, Zhaoyang and Ye, Shenglong and Gu, Lixin and Tian, Hao and Duan, Yuchen and Su, Weijie and Shao, Jie and others},
  journal={arXiv preprint arXiv:2504.10479},
  year={2025}
}

@article{rafailov2023direct,
  title={Direct preference optimization: Your language model is secretly a reward model},
  author={Rafailov, Rafael and Sharma, Archit and Mitchell, Eric and Manning, Christopher D and Ermon, Stefano and Finn, Chelsea},
  journal={Advances in neural information processing systems},
  volume={36},
  pages={53728--53741},
  year={2023}
}

@inproceedings{agrawal2018dont,
  title     = {Don't Just Assume; Look and Answer: Overcoming Priors for Visual Question Answering},
  author    = {Agrawal, Aishwarya and Batra, Dhruv and Parikh, Devi and Kembhavi, Aniruddha},
  booktitle = {Proceedings of the IEEE Conference on Computer Vision and Pattern Recognition},
  pages     = {4971--4980},
  year      = {2018}
}

@article{cadene2019rubi,
  title={Rubi: Reducing unimodal biases for visual question answering},
  author={Cadene, Remi and Dancette, Corentin and Cord, Matthieu and Parikh, Devi and others},
  journal={Advances in neural information processing systems},
  volume={32},
  year={2019}
}

@inproceedings{kv2020reducing,
  title={Reducing language biases in visual question answering with visually-grounded question encoder},
  author={Kv, Gouthaman and Mittal, Anurag},
  booktitle={European Conference on Computer Vision},
  pages={18--34},
  year={2020},
  organization={Springer}
}

@inproceedings{chen2022rich,
  title={Rich knowledge sources bring complex knowledge conflicts: Recalibrating models to reflect conflicting evidence},
  author={Chen, Hung-Ting and Zhang, Michael and Choi, Eunsol},
  booktitle={Proceedings of the 2022 Conference on Empirical Methods in Natural Language Processing},
  pages={2292--2307},
  year={2022}
}

@article{kortukov2024studying,
  title        = {Studying Large Language Model Behaviors Under Context-Memory Conflicts With Real Documents},
  author       = {Kortukov, Evgenii and Rubinstein, Alexander and Nguyen, Elisa and Oh, Seong Joon},
  journal      = {arXiv preprint arXiv:2404.16032},
  year         = {2024}
}

@inproceedings{zhao2021calibrate,
  title     = {Calibrate Before Use: Improving Few-Shot Performance of Language Models},
  author    = {Zhao, Tony Z. and Wallace, Eric and Feng, Shi and Klein, Dan and Singh, Sameer},
  booktitle = {Proceedings of the 38th International Conference on Machine Learning},
  pages     = {12697--12706},
  year      = {2021},
  volume    = {139},
  series    = {Proceedings of Machine Learning Research},
  publisher = {PMLR}
}

@article{han2022prototypical,
  title={Prototypical calibration for few-shot learning of language models},
  author={Han, Zhixiong and Hao, Yaru and Dong, Li and Sun, Yutao and Wei, Furu},
  journal={arXiv preprint arXiv:2205.10183},
  year={2022}
}

@article{nie2022improving,
  title   = {Improving Few-Shot Performance of Language Models via Nearest Neighbor Calibration},
  author  = {Nie, Feng and Chen, Meixi and Zhang, Zhirui and Cheng, Xu},
  journal = {arXiv preprint arXiv:2212.02216},
  year    = {2022}
}

@inproceedings{zhou2024batch,
  title={Batch calibration: Rethinking calibration for in-context learning and prompt engineering},
  author={Zhou, Han and Wan, Xingchen and Proleev, Lev and Mincu, Diana and Chen, Jilin and Heller, Katherine and Roy, Subhrajit},
  booktitle={International Conference on Learning Representations},
  volume={2024},
  pages={49--70},
  year={2024}
}

@article{brown2020language,
  title={Language models are few-shot learners},
  author={Brown, Tom and Mann, Benjamin and Ryder, Nick and Subbiah, Melanie and Kaplan, Jared D and Dhariwal, Prafulla and Neelakantan, Arvind and Shyam, Pranav and Sastry, Girish and Askell, Amanda and others},
  journal={Advances in neural information processing systems},
  volume={33},
  pages={1877--1901},
  year={2020}
}

@inproceedings{liu2022makes,
  title     = {What Makes Good In-Context Examples for {GPT}-3?},
  author    = {Liu, Jiachang and Shen, Dinghan and Zhang, Yizhe and Dolan, Bill and Carin, Lawrence and Chen, Weizhu},
  booktitle = {Proceedings of Deep Learning Inside Out: The 3rd Workshop on Knowledge Extraction and Integration for Deep Learning Architectures},
  pages     = {100--114},
  year      = {2022},
  publisher = {Association for Computational Linguistics}
}

@article{platt1999probabilistic,
  title={Probabilistic outputs for support vector machines and comparisons to regularized likelihood methods},
  author={Platt, John and others},
  journal={Advances in large margin classifiers},
  volume={10},
  number={3},
  pages={61--74},
  year={1999},
  publisher={Cambridge, MA}
}

@inproceedings{guo2017calibration,
  title={On Calibration of Modern Neural Networks},
  author={Guo, Chuan and Pleiss, Geoff and Sun, Yu and Weinberger, Kilian Q.},
  booktitle={Proceedings of the 34th International Conference on Machine Learning},
  pages={1321--1330},
  year={2017},
  volume={70},
  series={Proceedings of Machine Learning Research},
  publisher={PMLR}
}

@inproceedings{wang2024mitigating,
  title={Mitigating hallucinations in large vision-language models with instruction contrastive decoding},
  author={Wang, Xintong and Pan, Jingheng and Ding, Liang and Biemann, Chris},
  booktitle={Findings of the Association for Computational Linguistics: ACL 2024},
  pages={15840--15853},
  year={2024}
}

@article{park2024mitigating,
  title={Mitigating dialogue hallucination for large vision language models via adversarial instruction tuning},
  author={Park, Dongmin and Qian, Zhaofang and Han, Guangxing and Lim, Ser-Nam},
  journal={arXiv preprint arXiv:2403.10492},
  year={2024}
}

@article{chen2024we,
  title={Are we on the right way for evaluating large vision-language models?},
  author={Chen, Lin and Li, Jinsong and Dong, Xiaoyi and Zhang, Pan and Zang, Yuhang and Chen, Zehui and Duan, Haodong and Wang, Jiaqi and Qiao, Yu and Lin, Dahua and others},
  journal={Advances in Neural Information Processing Systems},
  volume={37},
  pages={27056--27087},
  year={2024}
}

@inproceedings{liu2024mmbench,
  title={Mmbench: Is your multi-modal model an all-around player?},
  author={Liu, Yuan and Duan, Haodong and Zhang, Yuanhan and Li, Bo and Zhang, Songyang and Zhao, Wangbo and Yuan, Yike and Wang, Jiaqi and He, Conghui and Liu, Ziwei and others},
  booktitle={European conference on computer vision},
  pages={216--233},
  year={2024},
  organization={Springer}
}

@inproceedings{deng2025words,
  title={Words or vision: Do vision-language models have blind faith in text?},
  author={Deng, Ailin and Cao, Tri and Chen, Zhirui and Hooi, Bryan},
  booktitle={2025 IEEE/CVF Conference on Computer Vision and Pattern Recognition (CVPR)},
  pages={3867--3876},
  year={2025},
  organization={IEEE}
}

@inproceedings{schwenk2022okvqa,
  title={A-okvqa: A benchmark for visual question answering using world knowledge},
  author={Schwenk, Dustin and Khandelwal, Apoorv and Clark, Christopher and Marino, Kenneth and Mottaghi, Roozbeh},
  booktitle={European conference on computer vision},
  pages={146--162},
  year={2022},
  organization={Springer}
}

@inproceedings{das2017visual,
  title={Visual dialog},
  author={Das, Abhishek and Kottur, Satwik and Gupta, Khushi and Singh, Avi and Yadav, Deshraj and Moura, Jos{\'e} MF and Parikh, Devi and Batra, Dhruv},
  booktitle={Proceedings of the IEEE conference on computer vision and pattern recognition},
  pages={326--335},
  year={2017}
}

\appendix

\section{Experimental Setup and Implementation Details}
\label{app:experimental_setup_details}
\label{app:dataset_construction}

This appendix provides implementation details and supplementary analyses that support the main text. 
We organize the material into experimental protocols, robustness checks, full results, and formulation discussions.

\subsection{Decision-Margin Computation Details}
\label{app:decision_margin_details}

All margin analyses in this paper use the same scoring convention.
Given a rendered multimodal input $x=(I,T)$, where $I$ is the image and $T$ is the complete text prompt, we define the binary decision margin as
\[
m(x)=\log P_\theta(v_{+}\mid I,T)-\log P_\theta(v_{-}\mid I,T),
\]
where $v_{+}$ and $v_{-}$ are the affirmative and negative verbalizers.
In the main experiments, $v_{+}=\texttt{Yes}$ and $v_{-}=\texttt{No}$.

We score both candidate continuations under teacher forcing.
For a candidate verbalizer $v=(v_1,\ldots,v_K)$ tokenized into $K$ tokens, its sequence log-probability is computed as
\[
\log P_\theta(v\mid I,T)
=
\sum_{k=1}^{K}
\log P_\theta(v_k\mid I,T,v_{<k}).
\]
We then subtract the sequence log-probabilities of the two verbalizers.
This sequence-level scoring is used even when a verbalizer consists of more than one token, so the reported margin is not restricted to a single next-token logit difference.

The final binary prediction induced by the margin is determined by its sign:
\[
\hat{y}(x)=
\begin{cases}
\texttt{Yes}, & m(x)>0,\\
\texttt{No}, & m(x)\le 0.
\end{cases}
\]

\subsection{Irrelevant Text Sampling}
\label{app:irrelevant_text_sampling}

We construct image-irrelevant textual contexts from an offline WikiText sentence pool. 
The contexts are not dynamically generated during inference; instead, each sample is assigned a fixed context before evaluation. 
This makes the context assignment deterministic and reproducible under the same random seed.

We first split corpus into sentence-level candidates and remove duplicate sentences. 
The candidate pool is then shuffled with a fixed seed, and contexts are assigned to samples by iterating through this shuffled pool. 

The filtering process consists of two steps. 
First, each candidate context is processed by a rule-based script to remove judgment-related words, such as ``yes'' and ``no''. 
Second, we compute the CLIP similarity between the candidate context and the input image, and discard candidates with high similarity scores to avoid selecting visually related sentences.

For CLIP filter, we calculate the similarity score between context and image:
\begin{equation}
    \mathrm{sim}(i, t) = \langle \mathrm{CLIP}(i), \mathrm{CLIP}(t) \rangle,
\end{equation}
where $i$ is the input image and $t$ is a candidate context. 

\subsection{More Reproducibility Details}

\paragraph{Reproducibility details.}
We use WikiText-103-v1 as the source of irrelevant context, with context lengths ranging from 13 to 30 words and an average length of 20.8 words. Context relevance is filtered using CLIP, with a similarity threshold of 0.14. Unless otherwise specified, all random seeds are fixed to 42. All evaluated model checkpoints are downloaded from Hugging Face. Experiments are conducted on NVIDIA RTX 3090 GPUs with CUDA 12.4. 



\paragraph{Agreement with greedy decoding.}
Our margin-based prediction and the parsed greedy-decoding output correspond to two distinct operational definitions of model prediction. We therefore measure their agreement by computing the proportion of samples for which the sign of the teacher-forced decision margin matches the parsed greedy-decoding answer. Across all evaluated datasets and models, the agreement exceeds 90\%, reaching up to 99\%. This high consistency indicates that the margin-level analysis closely reflects the model's actual decoded decisions rather than an artifact of teacher-forced scoring.

\begin{figure*}[t]
\centering
\includegraphics[width=\textwidth]{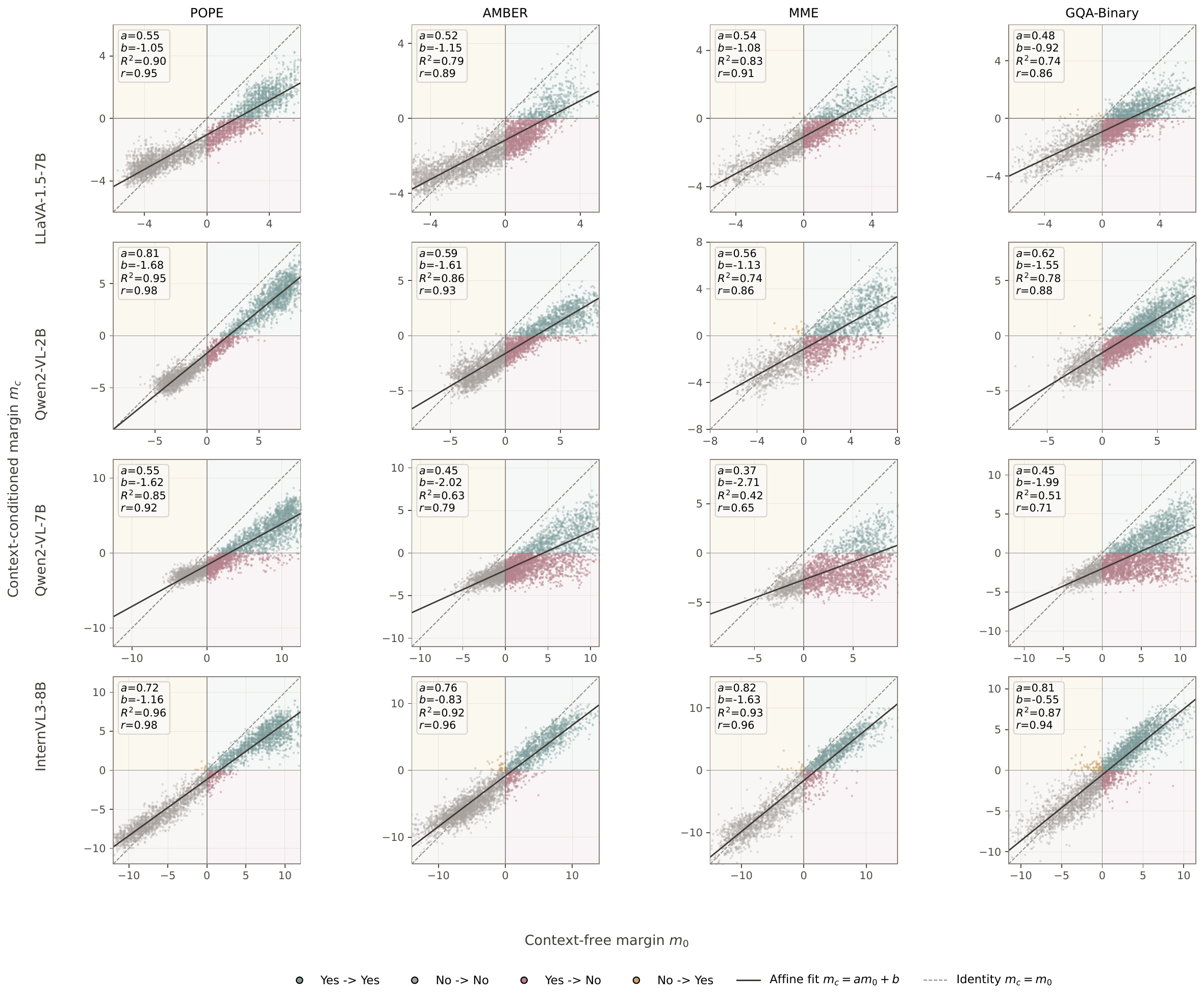}
\caption{
The affine transformation results across all models and benchmarks.
}
\label{fig:all}
\end{figure*}

\subsection{Benchmark Statistics and Implementation Details}
\label{app:dataset_statistics}

In the main experiments, we use four benchmarks. 
POPE, MME, and AMBER-discriminative are already binary judgment tasks, so we preserve their original question format and only add a unified answer instruction. 
POPE contains 8,910 questions, including 4,500 Yes and 4,410 No instances (50.5\%/49.5\%); its three evaluation categories contain 3,000 adversarial, 3,000 popular, and 2,910 random questions.  
The AMBER discriminative subset contains 14,216 questions, with 4,789 Yes and 9,427 No instances (33.7\%/66.3\%), and is therefore the
most label-imbalanced benchmark in our main evaluation.  The MME input contains 2,374 questions, evenly split into 1,187 Yes and 1,187 No instances across its perception and cognition subtasks.  

For GQA, we further divide the benchmark into two subsets: GQA-Binary and GQA-Judgment. 
GQA-Binary contains the native binary questions in GQA, for which we preserve the original question format and only append the unified answer instruction. It contains 4,525 naturally yes/no questions, with 2,328 Yes and 2,197 No instances (51.4\%/48.6\%). 
GQA-Judgment is constructed from the remaining open-ended questions by converting each sample into an answer-verification task: given the image, the original question, and a candidate answer, the model is asked to judge whether the candidate answer is supported by the image. 
See Appendix~\ref{app:gqa-judgment-results} for detailed conversion process and corresponding experiment results.

Our experiments show that this conversion changes the task format from the model's perspective. 
Although GQA-Judgment still exhibits the affine margin pattern observed in our main analysis, its fitted affine parameters differ from those of native binary tasks. 
Therefore, we report GQA-Binary in the main text for consistency with the primary binary-judgment setting, and provide GQA-Judgment results as a complementary evaluation.

\begin{table*}[t]
\centering
\tiny
\caption{
Absolute behavioral metrics across tested models and benchmarks.
Free denotes context-free input, Neutral denotes WikiText-after-neutral, and Condition denotes WikiText-after-possibly.
Part (b) reports the original-style MME score.
}
\label{tab:app_absolute_behavioral_metrics}
\resizebox{\textwidth}{!}{%
\begin{tabular}{llrrrrrrrrrrrr}
\toprule
\multirow{2}{*}{Model} & \multirow{2}{*}{Dataset}
& \multicolumn{4}{c}{Free}
& \multicolumn{4}{c}{Neutral}
& \multicolumn{4}{c}{Condition} \\
\cmidrule(lr){3-6}\cmidrule(lr){7-10}\cmidrule(lr){11-14}
& & Acc & F1 & Rec. & Prec. & Acc & F1 & Rec. & Prec. & Acc & F1 & Rec. & Prec. \\
\midrule
LLaVA-1.5-7B & POPE       & 85.5 & 84.2 & 76.6 & 93.5 & 84.8 & 83.2 & 74.2 & 94.6 & 85.2 & 83.8 & 75.7 & 93.8 \\
LLaVA-1.5-7B & AMBER      & 81.2 & 69.8 & 64.8 & 75.8 & 80.9 & 68.7 & 62.4 & 76.5 & 79.9 & 69.2 & 67.1 & 71.4 \\
LLaVA-1.5-7B & MME        & 78.5 & 78.9 & 80.5 & 77.4 & 78.6 & 77.2 & 72.6 & 82.5 & 77.6 & 76.3 & 72.1 & 81.0 \\
LLaVA-1.5-7B & GQA-Binary & 76.4 & 76.9 & 76.4 & 77.5 & 76.4 & 77.4 & 78.6 & 76.3 & 74.5 & 77.1 & 83.4 & 71.7 \\
\midrule
Qwen2-VL-2B & POPE       & 87.8 & 86.8 & 79.6 & 95.5 & 86.0 & 84.5 & 75.1 & 96.4 & 85.8 & 84.1 & 74.5 & 96.7 \\
Qwen2-VL-2B & AMBER      & 85.3 & 78.3 & 78.5 & 78.1 & 84.6 & 75.1 & 68.7 & 82.7 & 80.4 & 66.2 & 56.9 & 79.1 \\
Qwen2-VL-2B & MME        & 79.9 & 80.1 & 80.5 & 79.6 & 69.3 & 64.6 & 55.9 & 76.4 & 59.1 & 49.3 & 39.7 & 65.0 \\
Qwen2-VL-2B & GQA-Binary & 79.2 & 79.0 & 76.1 & 82.1 & 72.9 & 68.0 & 56.0 & 86.6 & 73.0 & 68.7 & 57.7 & 84.9 \\
\midrule
Qwen2-VL-7B & POPE       & 88.3 & 87.5 & 81.1 & 95.0 & 87.0 & 85.8 & 77.7 & 95.7 & 79.9 & 75.5 & 61.3 & 98.4 \\
Qwen2-VL-7B & AMBER      & 86.9 & 80.7 & 81.2 & 80.2 & 87.4 & 80.9 & 79.2 & 82.7 & 77.5 & 53.3 & 38.1 & 88.8 \\
Qwen2-VL-7B & MME        & 87.6 & 88.0 & 90.6 & 85.5 & 87.8 & 87.8 & 88.0 & 87.7 & 60.1 & 34.6 & 21.1 & 95.4 \\
Qwen2-VL-7B & GQA-Binary & 82.2 & 82.0 & 78.7 & 85.5 & 80.8 & 80.3 & 75.9 & 85.2 & 62.4 & 45.4 & 30.4 & 89.6 \\
\midrule
InternVL3 & POPE       & 90.9 & 91.0 & 90.7 & 91.3 & 90.1 & 89.7 & 85.4 & 94.4 & 90.6 & 90.3 & 86.8 & 94.1 \\
InternVL3 & AMBER      & 88.3 & 83.1 & 85.7 & 80.6 & 88.8 & 82.9 & 81.0 & 84.9 & 88.4 & 82.2 & 79.7 & 84.9 \\
InternVL3 & MME        & 87.9 & 87.6 & 85.3 & 90.0 & 84.8 & 83.2 & 75.4 & 92.7 & 84.2 & 82.6 & 74.8 & 92.1 \\
InternVL3 & GQA-Binary & 81.9 & 82.9 & 85.1 & 80.8 & 80.4 & 80.9 & 80.8 & 81.0 & 79.9 & 80.2 & 79.0 & 81.4 \\
\bottomrule
\end{tabular}
}
\end{table*}

\section{Supplementary Experiment Results}
\label{app:additional_results}

\subsection{Absolute Behavioral Metrics}
\label{app:absolute_behavioral_metrics}

Table~\ref{tab:app_absolute_behavioral_metrics} reports the absolute behavioral metrics for all tested model families.
Neutral uses WikiText-after-neutral, while Condition uses WikiText-after-possibly.
All binary metrics are percentages.

\subsection{Full Affine Transformation Results}
\label{app:affine_transformation_results}

Figure~\ref{fig:all} shows the affine transformation results across all models and benchmarks.

\subsection{Calibration Experiments}
\label{app:calibration}

\paragraph{In-domain calibration.}
First, we test whether affine calibration can recover context-free behavior when
the affine parameters are estimated and evaluated on held-out splits from the
same benchmark. Specifically, for each model and each dataset, we split the data
into a calibration split and a test split. The parameters $\hat{a}$ and
$\hat{b}$ are fitted on the calibration split of the same dataset, and then
applied to the corresponding test split.

\paragraph{Cross-dataset transfer calibration.}
Second, we evaluate a stronger transfer setting. Instead of fitting affine
parameters on the target dataset, we fit $\hat{a}$ and $\hat{b}$ on one source
dataset and directly apply them to other target datasets:
\begin{equation}
    m_{\mathrm{cal}}^{s \rightarrow t}
    =
    \frac{m_{c}^{t} - \hat{b}_{s}}{\hat{a}_{s}},
\end{equation}
where $\hat{a}_{s}$ and $\hat{b}_{s}$ are estimated from the source dataset
$s$, and $m_c^{t}$ denotes the context-conditioned margin on the target dataset
$t$. This setting uses no target-dataset calibration samples.

Table~\ref{tab:calibration_in_domain} reports the full in-domain calibration on
LLaVA-1.5-7B. Inverse-affine calibration consistently recovers a substantial
part of the context-induced degradation: it raises accuracy relative to
Context by 3.2--7.9 points and reduces flips by 10.5--21.5 points across all
four benchmarks. This shows that the fitted affine distortion remains
predictive on held-out samples. 

\begin{table*}[t]
\centering
\tiny
\caption{
In-domain post-hoc affine calibration under WikiText-after-possibly context.
Affine parameters are fitted on calibration splits and evaluated on held-out
test splits. Values are percentages averaged over five splits. Flip is relative
to Origin predictions.
}
\label{tab:calibration_in_domain}
\begin{tabular}{llrrrrrrrr}
\toprule
Model & Dataset
& Origin Acc. & Context Acc. & Cal. Acc.
& Origin F1 & Context F1 & Cal. F1
& Context Flip & Cal. Flip \\
\midrule
LLaVA-1.5-7B & POPE       & 86.1 & 78.0 & 84.6 & 85.6 & 72.6 & 83.2 & 16.6 &  6.1 \\
LLaVA-1.5-7B & AMBER      & 82.7 & 74.9 & 79.7 & 77.5 & 47.3 & 71.0 & 29.2 & 13.6 \\
LLaVA-1.5-7B & MME        & 78.7 & 68.5 & 76.5 & 79.2 & 56.5 & 76.4 & 30.2 & 12.6 \\
LLaVA-1.5-7B & GQA-Binary & 73.2 & 69.2 & 72.3 & 77.7 & 63.5 & 75.8 & 36.3 & 14.8 \\
\bottomrule
\end{tabular}
\end{table*}

Table~\ref{tab:calibration_transfer} reports cross-dataset transfer for
LLaVA-1.5-7B. In every source--target pair, transferred parameters improve
accuracy relative to the uncalibrated Context condition and sharply reduce
flips. In particular, POPE-fitted parameters improve target accuracy by 2.4
points on GQA-Binary and by over 5 points on AMBER and MME, without using any
target calibration examples. These results support the interpretation that,
for LLaVA, irrelevant context induces a substantial model-level distortion
rather than an unrelated benchmark-specific artifact.

\begin{table*}[t]
\centering
\small
\caption{
Cross-dataset affine calibration transfer on LLaVA-1.5-7B under
WikiText-after-possibly context. Source-fitted parameters are applied directly
to the target benchmark without target calibration samples. Values are
percentages.
}
\label{tab:calibration_transfer}
\begin{tabular}{llrrrrrr}
\toprule
Source & Target
& Origin Acc. & Cal. Acc.
& Origin F1 & Cal. F1
& Origin Flip & Cal. Flip \\
\midrule
POPE       & AMBER      & 75.0 & 80.1 & 47.7 & 70.3 & 29.1 & 14.7 \\
POPE       & MME        & 68.5 & 76.6 & 56.4 & 76.2 & 30.4 & 12.8 \\
POPE       & GQA-Binary & 69.3 & 71.7 & 63.6 & 75.9 & 36.2 & 13.1 \\
\midrule
AMBER      & POPE       & 78.2 & 85.2 & 72.8 & 84.0 & 16.5 &  5.2 \\
AMBER      & MME        & 68.5 & 76.5 & 56.4 & 77.0 & 30.4 & 12.4 \\
AMBER      & GQA-Binary & 69.3 & 70.3 & 63.6 & 75.6 & 36.2 & 12.0 \\
\midrule
MME        & POPE       & 78.2 & 84.9 & 72.8 & 83.6 & 16.5 &  5.6 \\
MME        & AMBER      & 75.0 & 80.2 & 47.7 & 70.9 & 29.1 & 14.1 \\
MME        & GQA-Binary & 69.3 & 71.0 & 63.6 & 75.7 & 36.2 & 12.4 \\
\midrule
GQA-Binary & POPE       & 78.2 & 84.3 & 72.8 & 82.5 & 16.5 &  7.2 \\
GQA-Binary & AMBER      & 75.0 & 79.9 & 47.7 & 68.4 & 29.1 & 16.4 \\
GQA-Binary & MME        & 68.5 & 76.2 & 56.4 & 74.8 & 30.4 & 14.0 \\
\bottomrule
\end{tabular}
\end{table*}

Overall, these results show that for LLaVA, the context-induced margin
distortion is structured and partially transferable: inverse-affine correction
recovers context-free decisions on held-out data and under cross-dataset
transfer, while also consistently improving accuracy and F1. 

\begin{table*}[t]
\centering
\small
\caption{
Prompt, template, length, and source ablations on POPE with LLaVA-1.5-7B.
Acc, Recall, YesRate, and Flip are reported in percentages.
}
\label{tab:app_prompt_ablation}
\begin{tabular}{llrrrrrr}
\toprule
Group & Variant & $a$ & $b$ & $R^2$ & Acc & Recall & Flip \\
\midrule
Position/template
& Wiki before + possibly & 0.726 & -0.041 & 0.943 & 86.4 & 82.8 &  3.7 \\
& Wiki after + possibly  & 0.554 & -1.046 & 0.896 & 78.2 & 57.8 & 16.5 \\
& Wiki before + neutral  & 0.755 &  0.084 & 0.949 & 86.2 & 82.7 &  3.7 \\
& Wiki after + neutral   & 0.706 & -1.243 & 0.921 & 80.1 & 61.6 & 14.5 \\
\midrule
Template only
& After + possibly & 0.583 & -0.910 & 0.944 & 80.8 & 63.3 & 13.5 \\
& After + neutral  & 0.819 & -0.455 & 0.976 & 85.1 & 74.3 &  6.7 \\
\midrule
Length
& Short Wiki  & 0.699 & -0.000 & 0.940 & 86.3 & 82.4 & 4.1 \\
& Medium Wiki & 0.726 & -0.041 & 0.943 & 86.4 & 82.8 & 3.7 \\
& Long Wiki   & 0.743 & -0.150 & 0.934 & 86.3 & 82.2 & 4.2 \\
\midrule
Source/shuffle
& Shuffled Wiki       & 0.796 &  0.037 & 0.965 & 86.7 & 82.8 &  3.2 \\
& COCO caption before & 0.617 &  0.192 & 0.896 & 85.7 & 84.7 &  6.0 \\
& COCO caption after  & 0.475 & -1.204 & 0.892 & 74.2 & 49.7 & 20.7 \\
\bottomrule
\end{tabular}
\end{table*}

\subsection{Results beyond Binary Judgments}
\label{app:mcqa}

To further examine whether the observed affine tendency is restricted to this controlled setting, we extend our evaluation to multiple-choice question answering (MCQA), as well as more realistic multimodal QA, RAG, and multi-turn dialogue scenarios, to assess its applicability across different answer spaces and context construction mechanisms.

\subsubsection{Results on MCQA Benchmarks}

We first extend our evaluation to two multiple-choice question answering (MCQA) benchmarks, MMStar~\cite{chen2024we} and MMBench~\cite{liu2024mmbench}. 
For MCQA, we define the decision margin as the log-probability difference between the correct option and the highest-probability incorrect option:

\begin{equation}
m=\log P(y^*)-\max_{y_j\in\mathcal{Y},\,y_j\neq y^*}\log P(y_j),
\end{equation}

where $y^*$ denotes the correct answer.

The results in Table~\ref{tab:mcqa} show that irrelevant context also alters model predictions under the MCQA setting, while the context-free and context-conditioned margins continue to exhibit a clear affine tendency. For example, under the Conditioned setting, LLaVA and Qwen2-VL-7B achieve $R^2=0.727$ and $0.812$ on MMStar, respectively. On MMBench, the $R^2$ values of the four models range from $0.799$ to $0.940$. Meanwhile, irrelevant context generally leads to a certain degree of accuracy degradation. Overall, these results suggest that the affine structure observed in the previous sections is not specific to the binary Yes/No answer space, but remains observable in more general multiple-choice decisions.

\subsubsection{Results on Realistic Scenarios}

We further evaluate the observed pattern in three real-world scenarios. For realistic QA, we use the VQAv2 subset of Word-or-Vision~\cite{deng2025words}. The results show that irrelevant context exhibits a clear affine tendency. 
For RAG, we conduct experiments on A-OKVQA~\cite{schwenk2022okvqa}. Irrelevant retrieved context maintains a strong affine relation across all four models, while causing a modest performance degradation. 
For multi-turn dialogue, we evaluate VisDial v1.0~\cite{das2017visual} by comparing the no-history baseline with mismatched history. 
Mismatched history consistently reduces accuracy across models while still exhibiting a stable affine tendency. The results are shown in Table~\ref{tab:realistic}.

Overall, these results show that the affine tendency observed in the controlled setting remains observable in more realistic QA, RAG, and multi-turn dialogue scenarios, while its fit quality and parameters vary across models, tasks, and context types. This further supports viewing the affine relation as a broadly applicable first-order approximation rather than a universal law with fixed parameters across settings.

\begin{table*}[t]
\centering
\small
\setlength{\tabcolsep}{6pt}
\caption{Affine margin patterns on MCQA benchmarks. 
$\Delta$Acc denotes the accuracy change relative to the context-free condition.
Neutral and Conditioned correspond to the two context presentation settings used in our main experiments.}
\label{tab:mcqa}
\begin{tabular}{lllrrrrr}
\toprule
Dataset & Model & Context & $\Delta$Acc & $a$ & $b$ & $R^2$ & $r$ \\
\midrule

\multirow{8}{*}{MMStar}
& \multirow{2}{*}{LLaVA-1.5-7B}
& Neutral     & -1.00 & 0.73 & -0.05 & 0.85 & 0.92 \\
& & Conditioned & -3.20 & 0.58 & -0.14 & 0.73 & 0.85 \\

& \multirow{2}{*}{Qwen2-VL-2B}
& Neutral     & -2.00 & 0.95 & -0.02 & 0.93 & 0.97 \\
& & Conditioned & -3.00 & 0.88 & -0.09 & 0.87 & 0.93 \\

& \multirow{2}{*}{Qwen2-VL-7B}
& Neutral     & -0.27 & 0.93 & -0.05 & 0.96 & 0.98 \\
& & Conditioned & -3.20 & 0.75 & -0.15 & 0.81 & 0.90 \\

& \multirow{2}{*}{InternVL3-8B}
& Neutral     & -5.27 & 0.83 & -0.09 & 0.76 & 0.87 \\
& & Conditioned & -4.33 & 0.83 & -0.02 & 0.78 & 0.88 \\

\midrule

\multirow{8}{*}{MMBench}
& \multirow{2}{*}{LLaVA-1.5-7B}
& Neutral     & -0.99 & 0.74 & -0.01 & 0.90 & 0.95 \\
& & Conditioned & -2.86 & 0.60 & -0.09 & 0.81 & 0.90 \\

& \multirow{2}{*}{Qwen2-VL-2B}
& Neutral     & -0.29 & 0.96 &  0.01 & 0.97 & 0.99 \\
& & Conditioned & -0.52 & 0.91 & -0.07 & 0.94 & 0.97 \\

& \multirow{2}{*}{Qwen2-VL-7B}
& Neutral     &  0.00 & 0.88 &  0.08 & 0.97 & 0.99 \\
& & Conditioned & -1.58 & 0.77 & -0.08 & 0.88 & 0.94 \\

& \multirow{2}{*}{InternVL3-8B}
& Neutral     & -1.70 & 0.80 &  0.12 & 0.81 & 0.90 \\
& & Conditioned & -1.05 & 0.79 &  0.26 & 0.80 & 0.89 \\

\bottomrule
\end{tabular}
\end{table*}

\begin{table*}[t]
\centering
\small
\setlength{\tabcolsep}{5pt}
\caption{Affine margin patterns in more realistic contextual scenarios.
$\Delta$Acc denotes the accuracy change relative to the corresponding context-free or no-history condition.
A dash indicates that the metric was not reported for that setting.}
\label{tab:realistic}
\begin{tabular}{lllrrrrr}
\toprule
Scenario & Model & Context & $\Delta$Acc & $a$ & $b$ & $R^2$ & $r$ \\
\midrule

\multicolumn{8}{l}{\textit{Realistic QA: VQAv2 subset of Word-or-Vision}} \\

& \multirow{3}{*}{LLaVA-1.5-7B}
& Matched    & -- & 0.78 & -1.87 & 0.60 & 0.77 \\
& & Irrelevant & -- & 0.96 &  0.32 & 0.89 & 0.95 \\
& & Corrupted  & -- & 0.92 & -2.21 & 0.57 & 0.75 \\

& \multirow{3}{*}{Qwen2-VL-2B}
& Matched    & -- & 0.54 & -3.57 & 0.48 & 0.69 \\
& & Irrelevant & -- & 0.87 & -0.87 & 0.81 & 0.90 \\
& & Corrupted  & -- & 0.85 & -3.60 & 0.37 & 0.61 \\

& \multirow{3}{*}{Qwen2-VL-7B}
& Matched    & -- & 0.34 & -5.99 & 0.38 & 0.62 \\
& & Irrelevant & -- & 0.55 & -3.82 & 0.58 & 0.76 \\
& & Corrupted  & -- & 0.57 & -5.22 & 0.38 & 0.62 \\

& \multirow{3}{*}{InternVL3-8B}
& Matched    & -- & 0.96 &  0.74 & 0.73 & 0.85 \\
& & Irrelevant & -- & 0.93 & -0.59 & 0.92 & 0.96 \\
& & Corrupted  & -- & 0.53 & -1.95 & 0.36 & 0.60 \\

\midrule

\multicolumn{8}{l}{\textit{Multimodal RAG: A-OKVQA}} \\

& \multirow{2}{*}{LLaVA-1.5-7B}
& Irrelevant & -1.92 & 0.75 & -0.07 & 0.87 & 0.93 \\
& & Supporting & +8.47 & 0.84 & +1.14 & 0.84 & 0.92 \\

& \multirow{2}{*}{Qwen2-VL-2B}
& Irrelevant & -0.17 & 0.92 & -0.01 & 0.97 & 0.99 \\
& & Supporting & +8.91 & 0.76 & +1.33 & 0.85 & 0.92 \\

& \multirow{2}{*}{Qwen2-VL-7B}
& Irrelevant & -0.09 & 0.94 & +0.13 & 0.95 & 0.98 \\
& & Supporting & +8.73 & 0.77 & +2.28 & 0.83 & 0.91 \\

& \multirow{2}{*}{InternVL3-8B}
& Irrelevant & -0.70 & 0.87 & +0.31 & 0.86 & 0.93 \\
& & Supporting & +8.82 & 0.64 & +2.74 & 0.51 & 0.72 \\

\midrule

\multicolumn{8}{l}{\textit{Multi-turn Dialogue: VisDial v1.0}} \\

& \multirow{2}{*}{LLaVA-1.5-7B}
& Gold history       & -0.70 & 0.76 & +0.10 & 0.66 & 0.81 \\
& & Mismatched history & -4.40 & 0.83 & -0.12 & 0.75 & 0.87 \\

& \multirow{2}{*}{Qwen2-VL-2B}
& Gold history       & -3.72 & 0.72 & -0.03 & 0.69 & 0.83 \\
& & Mismatched history & -4.72 & 0.80 & -0.12 & 0.78 & 0.88 \\

& \multirow{2}{*}{Qwen2-VL-7B}
& Gold history       & +0.72 & 0.86 & +0.18 & 0.80 & 0.90 \\
& & Mismatched history & -1.38 & 0.82 & -0.05 & 0.85 & 0.92 \\

& \multirow{2}{*}{InternVL3-8B}
& Gold history       & +2.74 & 0.80 & +0.28 & 0.70 & 0.83 \\
& & Mismatched history & -2.82 & 0.81 & -0.02 & 0.75 & 0.87 \\

\bottomrule
\end{tabular}
\end{table*}

\section{Additional Empirical Analysis}
\label{app:robustness_ablations}

\subsection{Prompt, Template, and Context-Source Ablations}
\label{app:prompt_ablation}

We further test whether the affine effect depends on a particular prompt rendering choice.
All experiments in this subsection use POPE with LLaVA-1.5-7B and reuse the same context-free origin outputs.
We vary four factors: the position of the context relative to the question, the wording of the context template, the context length, and the context source.
We also include template-only controls, where the template phrase is inserted without any external context text.
Table~\ref{tab:app_prompt_ablation} shows that the affine structure is robust across prompt variants, but the fitted parameters are sensitive to how the context is introduced.

Putting WikiText before the question has a much weaker effect than putting it after the question.
For the possibly-related template, the before-question variant has $a=0.726$ and only $3.7\%$ flips, whereas the after-question variant has $a=0.554$ and $16.5\%$ flips.
However, in many practical in-context learning pipelines, especially settings where auxiliary textual information is appended as side information or retrieved evidence before answer generation, the added context is appended after the question. Our after-question setting is designed to reflect this practically relevant regime rather than to claim position-invariant behavior. 
This position ablation shows that the affine structure remains observable across position variants, therefore does not affect the conclusions in the main paper.

The template-only controls address the concern that the comparison between context-free and context-conditioned prompts may be confounded by the added template itself.
Even without external context text, the after-question template changes the margin distribution.
However, the neutral template-only setting is substantially weaker than the possibly-related template-only setting: its slope is closer to identity ($a=0.819$) and its flip rate is lower ($6.7\%$ vs. $13.5\%$).
Thus, template wording is not irrelevant, but it does not eliminate the need to model the full context-conditioned transformation.

Length and source controls provide additional checks.
Short, medium, and long WikiText contexts all preserve strong affine fits and similar behavioral metrics when placed before the question.
Shuffling WikiText also preserves a strong affine relation, suggesting that the effect does not require coherent evidence about the queried object.
In contrast, COCO captions placed after the question produce the strongest degradation among these controls, with $a=0.475$ and a flip rate of $20.7\%$.
Overall, these ablations support the main interpretation: irrelevant text systematically transforms the decision margin, and the strength of the transformation depends on the semantic role and placement of the added text.

\subsection{Label Verbalization}
\label{app:label_verbalization}

Our primary analyses define a binary decision margin using the verbalizers
\texttt{Yes} and \texttt{No}.  This raises a potential confound: the observed
affine relation might arise from the lexical properties of this particular
answer pair, rather than from a structured transformation of the model's
visual decision.  We therefore evaluate alternative answer verbalizations
with LLaVA-1.5-7B.  The image, question, label, and fixed WikiText context
assignment are kept unchanged; only the response instruction and the two
verbalizer sequences used for margin computation are replaced.  For an
affirmative/negative pair $(v_+,v_-)$, the semantic binary margin is
\[
    m_v =
    \log P(v_+ \mid I,Q,C) -
    \log P(v_- \mid I,Q,C).
\]
We evaluate the context-free input (Origin), WikiText introduced through a
neutral template (Neutral), and WikiText introduced through the
possibly-related template (Possibly).

Table~\ref{tab:app_label_verbalization_pope} first reports the POPE results.
In addition to the standard \texttt{Yes}/\texttt{No} format, we use
\texttt{True}/\texttt{False} as a natural semantic alternative. 

\begin{table*}[t]
\centering
\small
\caption{Label-verbalization controls on POPE with LLaVA-1.5-7B.
For each verbalizer pair, margins score the semantically affirmative token
against the negative token.  Behavioral metrics and Flip are percentages;
affine quantities are fitted between Origin and each context condition.}
\label{tab:app_label_verbalization_pope}
\begin{tabular}{llrrrrrrrr}
\toprule
Verbalizer & Condition & Acc & F1 & YesRate & Flip & $a$ & $b$ & $R^2$ & $r$ \\
\midrule
\multirow{3}{*}{\texttt{Yes}/\texttt{No}}
& Origin   & 86.4 & 86.1 & 47.1 & --   & --    & --     & --    & --    \\
& Neutral  & 73.5 & 64.7 & 24.7 & 22.4 & 0.481 & -1.229 & 0.854 & 0.924 \\
& Possibly & 75.3 & 68.0 & 26.8 & 20.4 & 0.384 & -0.871 & 0.836 & 0.914 \\
\midrule
\multirow{3}{*}{\texttt{True}/\texttt{False}}
& Origin   & 85.6 & 84.7 & 44.1 & --   & --    & --     & --    & --    \\
& Neutral  & 80.9 & 82.5 & 58.7 & 16.5 & 0.600 &  0.587 & 0.791 & 0.889 \\
& Possibly & 62.5 & 72.7 & 86.6 & 42.6 & 0.510 &  1.124 & 0.747 & 0.864 \\
\bottomrule
\end{tabular}
\end{table*}

To verify that the result is not restricted to POPE, we further evaluate the
natural \texttt{True}/\texttt{False} alternative on the AMBER.  As shown in
Table~\ref{tab:app_label_verbalization_amber}, affine compression remains
visible on this additional benchmark under both prompt templates.

\begin{table}[t]
\centering
\small
\caption{Natural verbalizer control on AMBER discriminative with
LLaVA-1.5-7B using \texttt{True}/\texttt{False}.  Behavioral metrics and
Flip are reported in percentages.}
\label{tab:app_label_verbalization_amber}
\begin{tabular}{lrrrrrrrr}
\toprule
Condition & Acc & Yes & Flip & $a$ & $b$ & $R^2$  \\
\midrule
Origin   & 77.4 & 26.9 & --   & --    & --    & --   \\
Neutral  & 68.8 & 58.7 & 32.9 & 0.58 & 0.61 & 0.66  \\
Possibly & 48.4 & 84.2 & 57.4 & 0.47 & 1.04 & 0.57  \\
\bottomrule
\end{tabular}
\end{table}

The natural-verbalizer results support the main conclusion.  With
\texttt{True}/\texttt{False}, the estimated slope is below one in every
condition: $a=0.600/0.510$ on POPE and $a=0.578/0.473$ on AMBER for the
Neutral/Possibly settings.  Thus, the margin-compression pattern is neither
specific to the literal \texttt{Yes}/\texttt{No} tokens nor confined to one
benchmark.  However, the intercept is not verbalizer invariant:
\texttt{Yes}/\texttt{No} yields negative shifts on POPE, while
\texttt{True}/\texttt{False} yields positive shifts and substantially higher
positive-answer rates, particularly under Possibly. 
This result suggests that the preference shift induced by irrelevant context is governed by a structured affine transformation of decision margins, which cannot be fully captured by output performance metrics alone.

\subsection{Why Supporting Context Still Compresses Margins}
\label{app:supporting_slope_explanation}

The context-role intervention in Sec.~4 introduces a seemingly
counter-intuitive result.  Since supporting context is constructed to be
consistent with the ground-truth answer, one might expect it to strengthen
the clean visual decision and therefore yield a slope larger than one.  This
is not what we observe.  Table~\ref{tab:app_context_role_slope} reports the
controlled intervention on POPE with LLaVA-1.5-7B, where the image,
question, context position, and rendering template are fixed while only the
semantic role of the inserted context is changed.

\begin{table}[t]
\centering
\small
\caption{Context-role intervention on POPE with LLaVA-1.5-7B.
$\Delta m_{+}$ and $\Delta m_{-}$ denote the mean shift $m_c-m_0$ for
ground-truth \texttt{Yes} and \texttt{No} samples, respectively.}
\label{tab:app_context_role_slope}
\begin{tabular}{lrrrrr}
\toprule
Context & $a$ & $b$ & $R^2$ & $\Delta m_{+}$ & $\Delta m_{-}$ \\
\midrule
Irrelevant  & 0.554 & -1.046 & 0.896 & -1.96 & 0.16 \\
Supporting  & 0.626 & -0.287 & 0.873 & -0.94 & 0.61 \\
Conflicting & 0.367 & -1.228 & 0.869 & -2.63 & 0.59 \\
\bottomrule
\end{tabular}
\end{table}

Supporting context is substantially less destructive than the alternatives:
it has the largest slope, the smallest negative intercept, and a much smaller
negative shift for ground-truth \texttt{Yes} samples than irrelevant or
conflicting context.  Nevertheless, its slope remains below one
($a=0.626$).  This does not contradict the usefulness of supporting evidence,
because the fitted slope measures the global dynamic range of the
context-conditioned margins relative to clean margins; it does not directly
measure whether a context is label-consistent for an individual sample.

In particular, the supporting intervention is label-conditioned: positive
questions receive a statement that the queried object is present, whereas
negative questions receive a statement that it is absent.  Its effect is
therefore more appropriately summarized as
\[
    m_c = \alpha m_0 + \delta(y) + \epsilon,
\]
where $\delta(y)$ is a label-dependent shift, rather than as a pure margin
amplifier $m_c=\gamma m_0$ with $\gamma>1$.  On POPE, the clean mean margins
for ground-truth \texttt{Yes} and \texttt{No} samples are $2.10$ and
$-2.76$, giving a mean class separation of $4.86$.  Under supporting
context, the corresponding means are $1.16$ and $-2.15$, reducing the
separation to $3.31$.  Supporting text preserves the decision direction
better than other context roles, while still narrowing the overall margin
range; an ordinary least-squares fit thus naturally yields $a<1$.

This compression is also consistent with how the context is presented to the
model.  The supporting statement is introduced as caption-like natural
language under a possibly-related template, rather than as an oracle answer.
Consequently, the model may treat it as uncertain evidence, reduce reliance
on the original visual signal, or imperfectly process the negated form used
for negative samples.  The latter possibility is reflected by
$\Delta m_{-}=0.61$: because positive margins favor \texttt{Yes}, a perfectly
effective supporting statement for a \texttt{No} example would shift the
margin in the negative direction, whereas the observed average shift is
positive.

We therefore interpret supporting context as reducing harmful
context-induced distortion rather than amplifying clean visual margins beyond
their original scale.  The ordering
\[
    a_{\mathrm{supporting}} >
    a_{\mathrm{irrelevant}} >
    a_{\mathrm{conflicting}}
\]
still shows that the semantic role of text systematically changes the affine
operator.  At the same time, $a_{\mathrm{supporting}}<1$ indicates that even
label-consistent context is integrated as imperfect prompt-conditioned
evidence, not as deterministic supervision.

\section{Affine Robustness Analysis}
\label{app:robustness_analysis}

\begin{table*}[t]
\centering
\small
\caption{
Context-only prior subtraction on POPE with LLaVA-1.5-7B.
We report the affine fit between $m_0$ and the prior-subtracted margin $m_{\mathrm{sub}}=m_c-m_{\mathrm{prior}}$.
Acc, F1, YesRate, and Flip are reported in percentages.
}
\label{tab:app_context_prior_subtraction}
\begin{tabular}{llrrrrrrrr}
\toprule
Context & Prior estimator & $a$ & $b$ & $R^2$ & $r$ & Acc & F1 & YesRate & Flip \\
\midrule
neutral & context-only blank
& 0.705 & -0.872 & 0.898 & 0.948 & 81.7 & 78.5 & 34.5 & 11.4 \\
neutral & question+context blank
& 0.700 &  1.894 & 0.904 & 0.951 & 79.7 & 82.4 & 64.9 & 19.1 \\
neutral & context-only same image
& 0.705 & -0.653 & 0.902 & 0.950 & 82.7 & 80.0 & 35.9 & 10.0 \\
\midrule
possibly & context-only blank
& 0.552 & -0.233 & 0.868 & 0.932 & 83.3 & 81.3 & 38.5 &  8.2 \\
possibly & question+context blank
& 0.549 &  1.599 & 0.872 & 0.934 & 75.5 & 79.7 & 70.0 & 24.3 \\
possibly & context-only same image
& 0.551 &  0.103 & 0.862 & 0.928 & 84.8 & 83.5 & 41.9 &  6.4 \\
\bottomrule
\end{tabular}
\end{table*}

\subsection{Context-Only Prior Subtraction}
\label{app:context_prior_subtraction}

Another alternative explanation is that irrelevant context only contributes an independent answer prior.
Under this view, the context-conditioned margin could be decomposed as the clean margin plus a context-only bias term.
To test this, we estimate a context prior margin $m_{\mathrm{prior}}$ by removing the original image-question evidence and then subtract it from the context-conditioned margin:
\[
m_{\mathrm{sub}} = m_c - m_{\mathrm{prior}} .
\]
We consider three prior estimators: context-only with a blank image, question-plus-context with a blank image, and context-only with the original image.

Table~\ref{tab:app_context_prior_subtraction} reports the POPE results with LLaVA-1.5-7B.
The context-only prior itself is almost uncorrelated with the clean margin: across variants, its fitted slope is close to zero and $R^2$ is nearly zero.
After subtracting this prior, the affine relation remains strong, and the slope remains substantially below one.
This is the key observation: subtracting an additive answer prior improves behavior, but it does not remove the margin compression effect.

\begin{table*}[t]
\centering
\small
\caption{
Nested margin-model comparison on POPE with LLaVA-1.5-7B.
The context variant is WikiText-after-neutral.
Fit metrics evaluate $m_c$ prediction on held-out samples.
Calibration metrics evaluate the corrected margin as a binary prediction.
}
\label{tab:app_nested_margin_models}
\begin{tabular}{lrrrrrr}
\toprule
Model & $R^2$ & MAE & Flip F1 & Calib Acc & Calib F1 & Calib Flip \\
\midrule
Identity   & 0.522 & 1.337 & 0.000 & 0.800 & 0.756 & 0.145 \\
Shift-only & 0.760 & 0.934 & 0.711 & 0.854 & 0.841 & 0.054 \\
Scale-only & 0.642 & 1.254 & 0.000 & 0.800 & 0.756 & 0.145 \\
Full affine & \textbf{0.920} & \textbf{0.518} & \textbf{0.849} & \textbf{0.855} & \textbf{0.842} & \textbf{0.050} \\
\bottomrule
\end{tabular}
\end{table*}

\begin{table*}[t]
\centering
\small
\caption{
Local margin analysis under WikiText-after-possibly.
For each model and dataset, we report the global affine fit and the flip rate within low-, mid-, and high-margin regions.
Flip rates are percentages.
}
\label{tab:app_local_margin_fit}
\begin{tabular}{llrrrrrr}
\toprule
Model & Dataset & Global $a$ & Global $R^2$ & Trim $R^2$ & Low Flip & Mid Flip & High Flip \\
\midrule
LLaVA & POPE       & 0.554 & 0.896 & 0.899 & 42.2 &  6.4 & 0.1 \\
LLaVA & AMBER      & 0.525 & 0.786 & 0.785 & 52.4 & 31.6 & 2.6 \\
LLaVA & MME        & 0.542 & 0.830 & 0.805 & 52.1 & 33.0 & 5.3 \\
LLaVA & GQA & 0.476 & 0.743 & 0.691 & 56.3 & 42.3 & 9.6 \\
\midrule
Qwen2-2B & POPE       & 0.557 & 0.859 & 0.841 & 42.6 & 15.4 & 3.1 \\
Qwen2-2B & AMBER      & 0.592 & 0.863 & 0.848 & 46.0 & 18.7 & 3.0 \\
Qwen2-2B & MME        & 0.561 & 0.740 & 0.730 & 42.8 & 21.7 & 5.9 \\
Qwen2-2B & GQA & 0.615 & 0.775 & 0.750 & 53.6 & 33.0 & 5.5 \\
\bottomrule
\end{tabular}
\end{table*}

For the possibly-related context, the uncorrected context margin has $a=0.554$, $b=-1.046$, and $R^2=0.896$.
Subtracting the context-only prior changes the intercept and improves behavioral metrics, but the slope remains around $0.55$.
Similarly, under the neutral context, the prior-subtracted slope remains around $0.70$.
Therefore, the irrelevant-context effect cannot be fully explained as a standalone Yes/No prior added to the decision margin.
Instead, it also changes the effective scaling of the original visual evidence, consistent with the affine interpretation in the main text.

\subsection{Nested Margin Models}
\label{app:nested_margin_models}

One possible concern of the affine pattern is that irrelevant context only induces a simpler transformation, such as a global Yes/No answer-prior shift or a uniform confidence scaling.
To verify this concern, we compare four nested margin models:
\[
\begin{aligned}
m_c &= m_0, \qquad m_c = m_0+b,\\
m_c &= a m_0, \qquad m_c = a m_0+b.
\end{aligned}
\]
These correspond to identity, shift-only, scale-only, and full affine transformations, respectively.
We fit each model on a calibration split and evaluate it on held-out samples.
In addition to held-out $R^2$ and MAE, we also evaluate whether each model predicts which samples will flip after adding context.

Table~\ref{tab:app_nested_margin_models} reports the available result on POPE with LLaVA-1.5-7B under the WikiText-after-neutral setting.
The full affine model provides the best held-out fit, reducing MAE from $1.34$ for identity and $0.93$ for shift-only to $0.52$, and increasing held-out $R^2$ to $0.92$.
It also gives the strongest flip prediction performance, with F1 increasing from $0.71$ for shift-only to $0.85$.
This supports the conclusion that the context effect is not merely a constant answer-direction shift, nor merely a confidence scaling effect.

For reference, the clean prediction accuracy on the same held-out split is $86.4\%$ with F1 $85.9\%$.
Adding context without correction reduces accuracy to $80.0\%$ and F1 to $75.6\%$, with a flip rate of $14.5\%$.
The full affine correction reduces the flip rate to $5.0\%$ and recovers most of the lost F1.
The fact that the full affine model substantially improves held-out margin prediction over both shift-only and scale-only baselines is the key diagnostic: irrelevant context changes both the scale of the original margin and its answer-direction offset.

\subsection{Local Fits Across Margin Regions}
\label{app:local_margin_fit}

The global affine relation could be inflated by the dynamic range of high-confidence samples.
To check this, we fit the affine relation separately within low-, mid-, and high-margin regions, where the regions are defined by quantiles of $|m_0|$ for each model--dataset pair.
We also repeat the global fit after trimming extreme margins.
This analysis serves two purposes.
First, it tests whether affine regularity remains visible after removing extreme margins.
Second, it quantifies where prediction flips occur.
Table~\ref{tab:app_local_margin_fit} shows two consistent patterns.
First, the affine fit remains strong after trimming the extreme $5$--$10\%$ of margins.
For example, LLaVA on POPE has global $R^2=0.896$ and trimmed $R^2=0.899$, while Qwen2-2B on AMBER has global $R^2=0.863$ and trimmed $R^2=0.848$.
Thus, the affine pattern is not solely an artifact of a few very large margins.
Second, flips are concentrated near the decision boundary.
Across both model families and all four benchmarks, low-margin samples have much higher flip rates than high-margin samples.
For LLaVA, the low-margin flip rate ranges from $42.2\%$ on POPE to $56.3\%$ on GQA-Binary, while the corresponding high-margin flip rates range only from $0.1\%$ to $9.6\%$.
Qwen2-2B shows the same pattern.
This supports the analysis in the main text: irrelevant context changes the margin distribution globally, but the final prediction changes primarily when the original decision is close to the boundary.
It also motivates gated calibration, where correction is focused on near-boundary examples instead of being applied uniformly to all samples.

\subsection{Random Seed Robustness}
\label{app:random_context_seed_results}

The main experiment assigns one fixed irrelevant WikiText context to each
POPE example.  To test whether the affine relation is an artifact of a
particular random assignment, we independently rebuild the WikiText
assignment with seeds 13 and 21, while keeping the model, samples, context
position, and possibly-related rendering template fixed.  We reuse the
context-free margins and rerun only the context-conditioned inference.

\begin{table}[t]
\centering
\small
\caption{Random-context-seed robustness on POPE with LLaVA-1.5-7B under
the WikiText-after-possibly condition.  Behavioral metrics and Flip are
reported in percentages.}
\label{tab:app_multiseed_context}
\begin{tabular}{lrrrrrrrr}
\toprule
Seed & $a$ & $b$ & $R^2$ & $r$ \\
\midrule
13 & 0.5541 & -1.0441 & 0.8968 & 0.9470 \\
21 & 0.5536 & -1.0492 & 0.8952 & 0.9461 \\
\midrule
Mean & 0.5539 & -1.0466 & 0.8960 & 0.9466 \\
Std. & 0.0003 &  0.0036 & 0.0012 & 0.0006 \\
\bottomrule
\end{tabular}
\end{table}

The estimated parameters are nearly invariant across the two additional
context assignments.  Moreover, they closely match the main seed-42 fit
($a=0.5538$, $b=-1.0465$, $R^2=0.8964$, and $r=0.9468$).
Thus, the reported affine transformation is not attributable to an
idiosyncratic WikiText sampling seed.

\begin{figure*}[t]
\centering
\includegraphics[width=\textwidth]{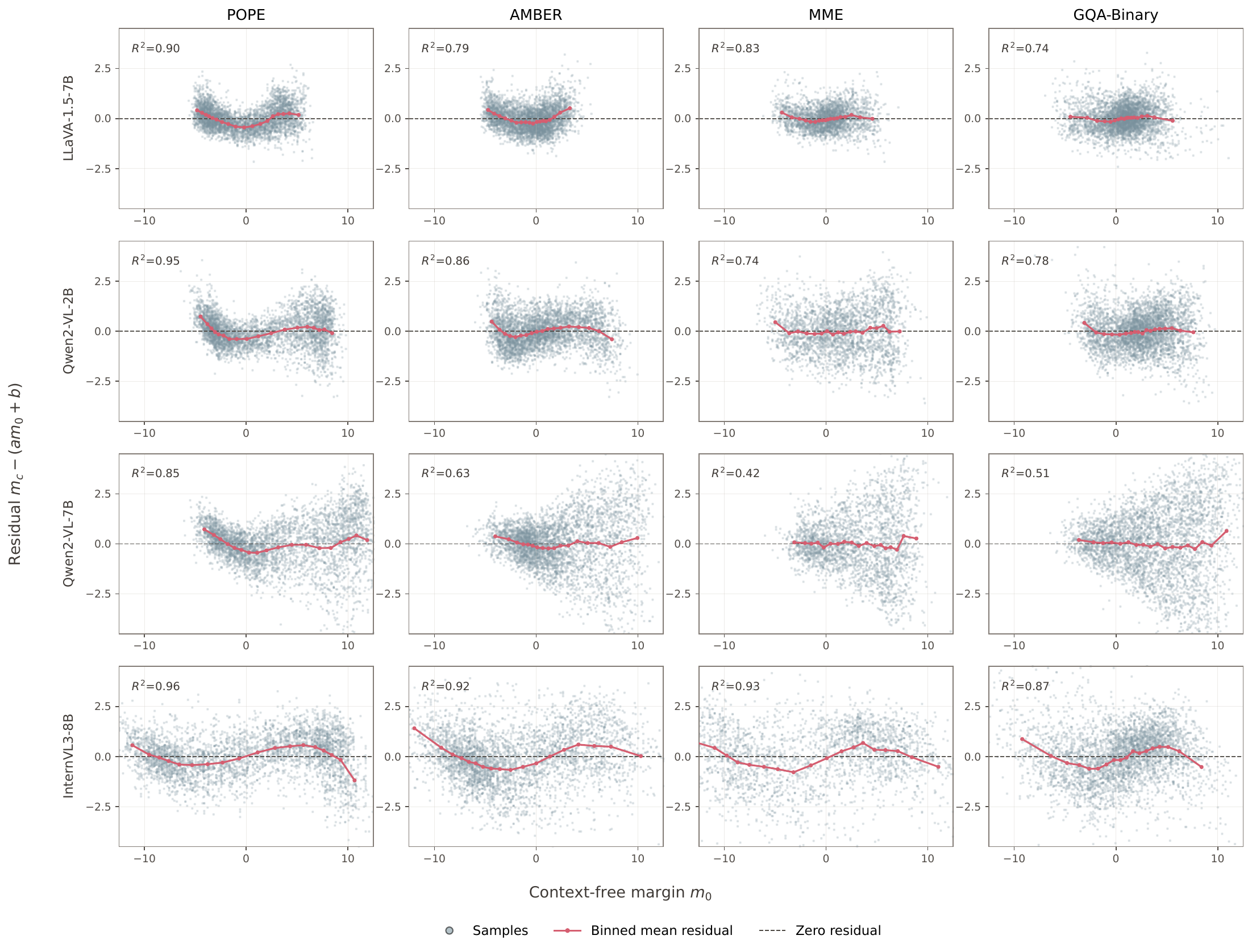}
\caption{Residual diagnostics for the affine fit under the
WikiText-after-possibly context.  Columns denote benchmarks and rows denote
models.  Each point is a sample residual
$m_c-(a m_0+b)$; the red curve is the binned mean residual.  All panels use
shared symmetric axes so that residual dispersion is directly comparable.}
\label{fig:app_residual_grid}
\end{figure*}

\subsection{Residual Analysis}
\label{app:residual_diagnostics}

We further examine whether the fitted affine transformation leaves systematic
structure in the prediction error.  For each model--benchmark pair under the
WikiText-after-possibly condition, we compute the residual
\begin{equation}
    \epsilon_i = m_c^{(i)} - \left(a m_0^{(i)} + b\right),
\end{equation}
where $a$ and $b$ are fitted on all samples of the corresponding setting.
Figure~\ref{fig:app_residual_grid} plots these residuals against the
context-free margin for four model families and four binary benchmarks.  The
red curve shows the mean residual within bins of $m_0$ and therefore exposes
systematic departures from a single global affine map.

For LLaVA-1.5-7B, the residual cloud remains comparatively compact across all
benchmarks, with $R^2$ ranging from $0.74$ on GQA-Binary to $0.90$ on POPE.
InternVL3-8B also retains high explanatory power ($R^2=0.87$--$0.96$),
although its larger margin scale produces a wider absolute residual spread.
Qwen2-VL-2B is broadly compatible with the affine approximation, with reduced
fit quality on MME and GQA-Binary.  In contrast, Qwen2-VL-7B exhibits visibly
larger and margin-dependent residual dispersion on AMBER, MME, and
GQA-Binary, where $R^2$ drops to $0.63$, $0.42$, and $0.51$, respectively.
Thus, the residual analysis supports the affine pattern as a strong first-order
description across models, while also identifying settings in which a single
global affine operator is incomplete.  This qualification is consistent with
the Qwen2-VL calibration results above: calibration can reduce context-induced
decision changes even when the remaining residual structure prevents uniform
accuracy recovery.

\section{Additional Discussion and Experiments}
\label{app:additional_discussion}

\subsection{Why Binary Judgment?}
\label{app:why_binary_judgment}

We adopt a binary judgment formulation to obtain a controlled and comparable measure of model preference. 
Open-ended multimodal generation involves many confounding factors, including answer phrasing, verbosity, decoding strategy, and automatic evaluation noise. 
These factors make it difficult to determine whether a change in the output reflects a genuine change in visual judgment or merely a change in surface generation. 
By constraining the answer space to $\{\texttt{Yes}, \texttt{No}\}$, we isolate the model's binary decision-making from the confounding effects of open-ended generation and uncontrolled answer formats.

This formulation also enables margin-level analysis. 
For each input, the model assigns probabilities to the same two candidate answers, allowing us to define a decision margin as the log-probability difference between \texttt{Yes} and \texttt{No}. 
Because the candidate set is fixed across the context-free and context-conditioned inputs, the margin provides a direct score-level probe of how the same image-question pair is affected by the added context. 
This is essential for identifying the affine relation between the original and context-conditioned margins.

The binary setting is not intended to replace open-ended evaluation. 
Rather, it serves as a controlled abstraction for studying whether image-irrelevant context changes visually grounded decisions. 
Several benchmarks, such as POPE and MME, are already naturally formulated as yes/no questions; for open-ended benchmarks, we convert them into answer-verification questions so that the same analysis can be applied consistently. 
This design allows us to compare predictions, flip rates, and margin shifts under a unified evaluation protocol.

\subsection{GQA-judgment Conversion and Results}
\label{app:gqa-judgment-results}

For non-binary GQA questions, Following~\cite{kalai2025language}, we convert them into answer-verification questions using scene-graph annotations. 
Given an original question and its ground-truth answer, we construct a positive verification sample by using the ground-truth answer as the candidate answer. 
To construct a negative sample, we sample an incorrect candidate from the GQA scene graph while preserving the semantic type of the answer whenever possible. 
The final question is converted into following format:
\begin{quote}
\small
\ttfamily
Question: [Question] Candidate answer: [Answer]. Is the candidate answer correct according to the image?
\end{quote}

\begin{table}[t]
\centering
\small
\caption{
GQA-Judgment behavior and affine fit for LLaVA under the WikiText-after-possibly context.
Origin accuracy is evaluated without irrelevant context, while context metrics are evaluated after inserting irrelevant context.
}
\label{tab:app_gqa_av_results}
\begin{tabular}{@{}lrlr@{}}
\toprule
\multicolumn{2}{c}{Affine Fit} & \multicolumn{2}{c}{Behavior} \\
\cmidrule(r){1-2} \cmidrule(l){3-4}
Metric & Value & Metric & Value \\
\midrule
Slope $a$       & 0.336  & Origin Acc.  & 81.3 \\
Intercept $b$   & -1.377 & Context Acc. & 61.6 \\
$R^2$           & 0.580  & Context F1   & 40.4 \\
Pearson $r$     & 0.762  & Yes Rate     & 14.5 \\
                &        & Flip Rate    & 32.1 \\
\bottomrule
\end{tabular}
\end{table}

Table~\ref{tab:app_gqa_av_results} shows that GQA-judgment behaves differently from the cleaner binary existence benchmarks. We use LLaVA as the testing model.
Without irrelevant context, the converted task is solvable: LLaVA reaches $81.3\%$ accuracy.
adding irrelevant context can strongly suppress affirmative verification judgments.
Under WikiText-after-possibly, the fitted slope is only $a=0.336$ with $R^2=0.580$, and the YesRate drops to $14.5\%$.
Additionally, the affine parameter is different, indicating that the prompt template or task will change the affine parameter.

\begin{figure}[t]
\centering
\includegraphics[width=0.5\textwidth]{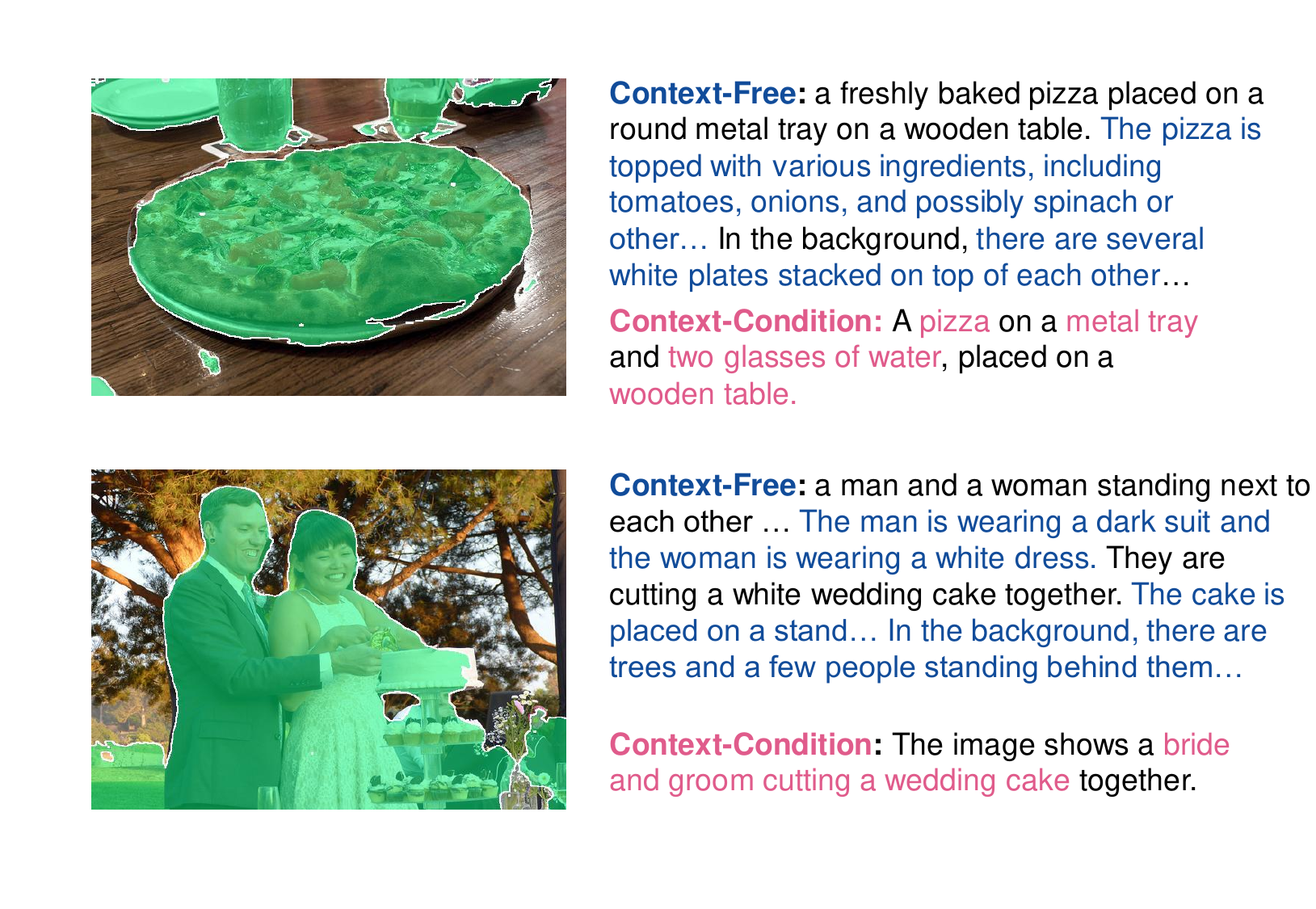}
\caption{
A case of our open-ended generation experiment. We use the green mask to indicate the region covered by the model's prediction under the context-condition setting. After adding irrelevant context, the model tends to describe only salient objects in the image, while peripheral objects are no longer mentioned. 
}
\label{fig:open}
\end{figure}

\subsection{Preliminary Study on Open-Ended Generation}
\label{app:open-ended}
To further examine whether irrelevant context affects open-ended generation, we conduct a preliminary study using AMBER generative set. 
We follow the setting of context-free and context-condition, and ask the model to describe each given image.
To be specific, for all settings, the question is:
\begin{quote}
\small
\ttfamily
Describe this image in detail.
\end{quote}

The \textbf{context-free} input contains only the image and the question:
\begin{quote}
\small
\ttfamily
[Image] [Question]
\end{quote}

The \textbf{context-condition} input appends the sampled irrelevant context and the template:
\begin{quote}
\small
\ttfamily
[Image] [Question] [Template] [Context]
\end{quote}

\begin{table}[t]
  \centering
  \small
  \setlength{\tabcolsep}{4.2pt}
  \caption{
  AMBER generative evaluation with LLaVA-1.5-7B.
  \texttt{Free} and \texttt{Condition} denote context-free and context-conditioned generation, respectively.
  Hal., Cov., Cog., and Len. denote hallucination rate, object coverage, cognitive error, and average output length.
  }
  \label{tab:app_amber_generative_context}
  \begin{tabular}{lrrrrr}
  \toprule
  Setting & CHAIR $\downarrow$ & Cov. $\uparrow$ & Hal. $\downarrow$ & Cog. $\downarrow$ & Len. \\
  \midrule
  \texttt{Free}      & 7.9 & 49.4 & 33.3 & 4.0 & 68.1 \\
  \texttt{Condition} & 3.8 & 39.3 &  9.5 & 0.7 & 16.1 \\
  \bottomrule
  \end{tabular}
\end{table}

The experiment result is in Table~\ref{tab:app_amber_generative_context}. 
We find that adding irrelevant context substantially changes the model's behavior: the generated responses become significantly shorter, accompanied by a large drop in coverage. 

The example in Figure~\ref{fig:open} provides an intuitive illustration of this behavioral shift: the model tends to describe only salient objects in the image, while peripheral objects are no longer mentioned. 

One possible interpretation is that irrelevant context changes the model's effective decision margin rather than simply injecting random noise into generation. Under the context-induced affine transformation, many visual claims are shifted toward a more conservative decision region. Consequently, objects or attributes that originally receive only weak or moderate support are more likely to fall below the implicit threshold for verbalization. These marginal claims often correspond to small objects, peripheral regions, fine-grained attributes, or visually ambiguous details. By contrast, highly salient objects tend to have stronger visual evidence and larger margins, making them more robust to the margin shift and more likely to remain in the output. This explains why the model still describes the dominant objects in the scene while omitting many peripheral details. 

From this perspective, the reduced coverage is not merely a side effect of shorter responses, but reflects a systematic change in the model's generation policy: irrelevant context makes the model more selective about which visual claims to express, suppressing uncertain details and producing more conservative descriptions.

However, further analysis is needed to verify the causal link between margin suppression and reduced coverage, and to better understand which visual claims are most affected.

\end{document}